\pdfoutput=1
\RequirePackage{fix-cm}
\documentclass[acmtog,british,screen=true]{acmart}
\acmSubmissionID{317}

\usepackage[utf8]{inputenc}
\usepackage[T1]{fontenc}
\usepackage[british]{babel}

\usepackage{pifont}
\newcommand{\cmark}{\ding{51}}
\newcommand{\xmark}{\ding{55}}

\newcommand{\tqdb}{\textquotedbl} % For straight quotation marks in tables

\usepackage{epsfig}
\usepackage{adjustbox} % Extends graphicx and \includegraphics
\graphicspath{{./figures/}}
\DeclareGraphicsExtensions{.pdf,.png,.jpeg,.jpg}

\usepackage{esint} % More integrals for computer modern math fonts
\usepackage{commath} % Provides \dif upright d
\usepackage{units} % Provides \nicefrac

\usepackage{booktabs} % For formal tables
\usepackage{array} % Patches and improves the array and tabular environments
\usepackage{rotating}
\usepackage{multirow}

\usepackage{enumitem} % Better control of lists and their spacing

\hypersetup{pdftitle={MoGlow: Probabilistic and controllable motion synthesis using normalising flows},%
 pdfauthor={Gustav Eje Henter, Simon Alexanderson, Jonas Beskow},%
 pdfkeywords={Generative models, machine learning, normalising flows, Glow, footstep analysis, data dropout}} % This information seems to somehow be overwritten by the template

\newcommand{\bb}[1]{\boldsymbol{#1}}

\newcommand{\given}[2]{\left(#1\,\middle|\,#2\right)}

\renewcommand{\mid}{\,\ifnum\currentgrouptype=16 \middle\fi|\,}

\newcommand{\real}{\mathbb{R}}

\newcommand{\Gauss}{\mathcal{N}}

\newcommand{\new}[1]{{#1}} % For hiding revisions
\newcommand{\newer}[1]{{#1}} % For hiding revisions

\newcommand{\man}{\textsuperscript{h}}
\newcommand{\dog}{\textsuperscript{d}}

\newcommand{\X}{\boldsymbol{X}}
\newcommand{\x}{\boldsymbol{x}}

\newcommand{\Z}{\boldsymbol{Z}}
\newcommand{\z}{\boldsymbol{z}}
\newcommand{\Ctrl}{\boldsymbol{C}}
\newcommand{\ctrl}{\boldsymbol{c}}
\newcommand{\f}{\boldsymbol{f}}

\newcommand{\Xrange}[2]{\bb{X}_{#1:#2}}
\newcommand{\xrange}[2]{\bb{x}_{#1:#2}}

\newcommand{\crange}[2]{\bb{c}_{#1:#2}}

\setlist{nolistsep}
\let\oldmarginpar\marginpar
\renewcommand\marginpar[1]{\-\oldmarginpar[\raggedleft\footnotesize #1]%
{\raggedright\footnotesize #1}}

\acmJournal{TOG}
\acmVolume{39}
\acmNumber{4}
\acmArticle{236}
\acmYear{2020}
\acmMonth{11}

\setcopyright{rightsretained}
\acmDOI{10.1145/3414685.3417836}

\begin{document}
% Title portion
\title[MoGlow: Probabilistic and Controllable Motion Synthesis Using Normalising Flows]{MoGlow: Probabilistic and Controllable Motion Synthesis Using~Normalising Flows}

% DO NOT ENTER AUTHOR INFORMATION FOR ANONYMOUS TECHNICAL PAPER SUBMISSIONS TO SIGGRAPH 2019!
\author{Gustav Eje Henter}
\authornote{Gustav Eje Henter and Simon Alexanderson contributed equally and are joint first authors.}
\orcid{0000-0002-1643-1054}
\email{ghe@kth.se}
\author{Simon Alexanderson}
\authornotemark[1]
\orcid{0000-0002-7801-7617}
\email{simonal@kth.se}
\author{Jonas Beskow}
\orcid{0000-0003-1399-6604}
\email{beskow@kth.se}
\affiliation{%
  \institution{Division of Speech, Music and Hearing, KTH Royal~Institute of Technology}
%  \streetaddress{104 Jamestown Rd}
  \city{Stockholm}
%  \state{VA}
%  \postcode{23185}
  \country{Sweden}}

\begin{abstract}
Data-driven modelling and synthesis of motion is an active research area with applications that include animation, games, and social robotics. This paper introduces a new class of probabilistic, generative, and controllable motion-data models based on normalising flows. Models of this kind can describe highly complex distributions, yet can be trained efficiently using exact maximum likelihood, unlike GANs or VAEs. Our proposed model is autoregressive and uses LSTMs to enable arbitrarily long time-dependencies. Importantly, is is also causal, meaning that each pose in the output sequence is generated without access to poses or control inputs from future time steps; this absence of algorithmic latency is important for interactive applications with real-time motion control. The approach can in principle be applied to any type of motion since it does not make restrictive, task-specific assumptions regarding the motion or the character morphology. We evaluate the models on motion-capture datasets of human and quadruped locomotion. Objective and subjective results show that randomly-sampled motion from the proposed method outperforms task-agnostic baselines and attains a motion quality close to recorded motion capture.
\end{abstract}




% A "teaser" image appears between the author and affiliation
% information and the body of the document, and typically spans the
% page.
\begin{teaserfigure}
%\vspace{-1ex} % For a 3.33:1 splash
%\vspace{-2ex} % For a 4.67:1 splash
\vspace{-1ex} % For a 5:1 splash
\includegraphics[width=\textwidth]{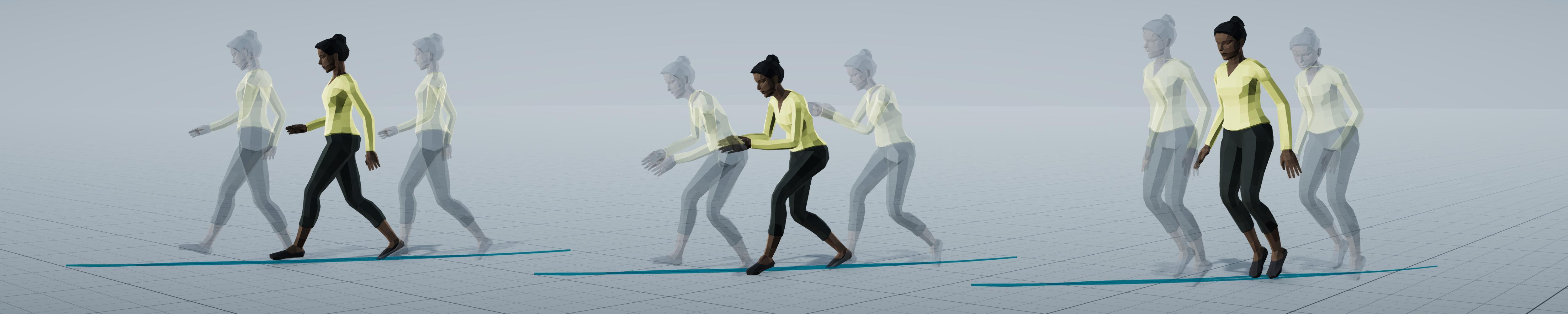}
%\vspace{-3.5ex} % For a 3.33:1 splash
%\vspace{-4ex} % For a 4.67:1 splash
\vspace{-4ex} % For a 5:1 splash
\caption{\new{Probabilistic motion generation.
Random samples from our method can give many distinct output motions even if the input signal is the same.}}
\label{fig:randomness}
%\vspace{1ex} % For a 3.33:1 splash
\vspace{1ex} % For a 5:1 splash
\Description{A visualisation of a character moving along the same path three times, but while one version of the character is walking normally, another is walking in a stance as if aiming a weapon, and the third is hopping instead of walking.}
\end{teaserfigure}

\maketitle

\section{Introduction}
\label{sec:introduction}
%\begin{figure}
%\centering
%\includegraphics[width=\columnwidth]{random_sampling}
%\vspace{-4ex}
%\caption{\new{Probabilistic motion generation.
%Random samples from our method can give many distinct output motions even if the input signal is the same.}}
%\label{fig:randomness}
%\vspace{-2.5ex}
%\Description{An image of three copies of a character moving along the same path, but while one version is walking normally, another is walking in a stance as if aiming a weapon, and the third is hopping instead of walking.}
%\end{figure}
A
%\blfootnote{\textsuperscript{*} Equal contribution.}
recurring problem in fields such as computer animation, video games, and artificial agents is how to generate convincing motion conditioned on high-level, ``weak'' control parameters.
Video-game characters, for example, should be able to display a wide range of motions controlled by game-pad inputs, and embodied agents should generate complex non-verbal behaviours based on, e.g., semantic and prosodic cues.
%While these types of problems have traditionally been solved by splicing together ``canned'' segments of motion capture data, 
%For many applications, the ability to generate naturalistic motion has an great impact on user experience and immersion.
%Consider, for example, a video-game character that is controlled in real time from a gamepad. Based on the control signals, the character should not only move realistically in different directions, but also change the style of locomotion between, e.g., walking and running as well as perform various actions such as jumping or dodging. This type of problem is traditionally solved by splicing together ``canned'' motion segments recorded using motion capture. Motion produced by such systems may look realistic at first sight, but humans observers rapidly catch on to the highly deterministic and repetitive nature of motion-capture playback, making it very hard and costly to build truly believable interactive motion with this methodology.
The advent of deep learning and the growing availability of large motion-capture databases have increased the interest in data-driven, statistical models for generating motion.
Given that the control signal is weak, a fundamental challenge for such models is to handle the large variation of possible outputs -- the limbs of a real person walking the same path twice will always follow different trajectories.
%that can differ in stride length and period 
%, instead of direct playback of motion clips. In general terms, 
%Furthermore, \emph{real-time} interactive systems require models with lowest possible latency, in  
%the ability to generate complex and naturalistic motion given only a \emph{weak control signal} (e.g., walk in direction $X$ at a pace of $Y$ m/s). It is important to note that there usually are many possible motion realisations that satisfy any given control signal -- the limbs of a real person who is asked to walk the same path twice, at the same speed, will always follow different trajectories.
%that can differ in stride length and period. Canned motion generally cannot replicate this.
%The weaker the control signal, the greater the range of motions that are consistent with that control signal. For example, ``walk forwards'' does not specify which leg moves first.
Deterministic models of motion, which return a single predicted motion,
%such as the estimated average pose for each time frame,
suffer from regression to the mean pose and produce artefacts like foot sliding in the case of gait.
\new{They also lack motion diversity, leading to repetitive and non-engaging characters in applications.}
Taken together, we are led to conclude that for motion generated from the model to be perceived as realistic, it \emph{cannot} be completely deterministic, but the model should instead generate \emph{different} motions upon each subsequent invocation, given the same control signal.
In other words, a stochastic model is required.
Furthermore, \emph{real-time} interactive systems such as video games require models with the lowest possible latency.

This paper introduces MoGlow, a novel autoregressive architecture for generating motion-data sequences based on normalising flows \cite{deco1994higher,dinh2015nice,dinh2017density,huang2018neural,kingma2018glow}.
This new modelling paradigm has the following principal advantages:
%This paper describes several novel models for generating motion-data sequences, all of which are based on normalising flows \cite{deco1994higher,dinh2015nice,dinh2017density,kingma2018glow}. This modelling paradigm has the following principal advantages:
\begin{enumerate}
\item It is \emph{probabilistic}, meaning that it %does not just describe one plausible motion, but 
endeavours to describe not just one motion, but \emph{all} possible motions, and how likely each possibility is. Plausible motion samples can then be generated also in the absence of conclusive control-signal input \new{(Fig.\ \ref{fig:randomness})}.
\item It uses an \emph{implicit model structure} \cite{mohamed2016learning} to parameterise distributions. This makes it fast to sample from without assuming that observed values follow restrictive, low-degree-of-freedom parametric families such as Gaussians or their mixtures, as done in, e.g., \citet{fragkiadaki2015recurrent,uria2015modelling}.
\item It allows exact and tractable probability computation, unlike variational autoencoders (VAEs) \cite{kingma2014auto,rezende2014stochastic}, and can be \emph{trained to maximise likelihood directly}, unlike generative adversarial networks (GANs) \cite{goodfellow2014generative,goodfellow2016nips}.
\item It is \emph{task-agnostic} -- that is, it does not rely on restrictive, situational assumptions such as characters being bipedal or motion being quasi-periodic (unlike, e.g., \citet{holden2017phase}).
\item It generates output
%autoregressively and
sequentially and permits control schemes for the output motion with \emph{no algorithmic latency}.
\item It is capable of generating \emph{high-quality motion} both in objective terms and as judged by human observers.
\end{enumerate}
To the best of our knowledge, our proposal is the first motion model based on normalising flows.
We evaluate our method on locomotion synthesis for two radically different morphologies -- humans and dogs -- since locomotion makes it easy to quantify artefacts and spot poor adherence to the control.
%The remainder of this paper is organised as follows: Sec.\ \ref{sec:background} describes related work in sequence modelling and motion synthesis.
%Sec.\ \ref{sec:method} then describes normalising flows and how we adapt them to describe motion sequences.
%Secs.\ \ref{sec:experiment} and \ref{sec:results} report on our experiments and their results, while Sec.\ \ref{sec:conclusion} concludes.
\href{https://youtu.be/pe-YTvavbtA}{A video presentation of our work} is available on YouTube, with \href{https://simonalexanderson.github.io/MoGlow/}{more information on our project page}.
%A video presentation of our work with generated motion examples
%can be found at \href{https://youtu.be/ozVldUcFjZg}{https://youtu.be/ozVldUcFjZg}.
%is enclosed in the supplement.
%\new{, along with video of results from a non-locomotion application}

\section{Background and prior work}
\label{sec:background}
%This section reviews recent developments in deep-learning-based generative models and introduce prior art in the domain of motion generation from motion-capture data.
\new{Mathematically, motion generation requires creating a sequence of poses from control input.
We here review (Sec.\ \ref{ssec:seqmodels}) probabilistic machine-learning models of sequences, and then describe (Secs.\ \ref{ssec:motionmodels} and \ref{ssec:probmotion}) prior work on machine learning for motion synthesis.}

\subsection{Probabilistic generative sequence models}
\label{ssec:seqmodels}
Probabilistic sequence models for continuous-valued data have a long history, with linear autoregressive models being an early example \cite{yule1927method}.
%These are simple models where inference (computing probabilities), parameter estimation, and sampling are fast and easy, at the expense of expressivity.
Model flexibility improved with the introduction of hidden-state models like HMMs \cite{rabiner1989tutorial} and Kalman filters \cite{welch1995introduction}, both of which still allow \new{efficient probability computation (\emph{inference})}.
Deep learning extended autoregressive models of continuous-valued data further by enabling highly nonlinear dependencies on previous observations, for example \citet{graves2013generating,zen2014deep,fragkiadaki2015recurrent,uria2015modelling}, as well as nonlinear (continuous-valued) hidden-state evolution through recurrent neural networks, e.g., \citet{hochreiter1997long}.
%For tractable inference, outputs nonetheless remain explicitly defined as Gaussians or mixture distributions.
%The same is true for nonparametric sequence models based on Gaussian processes \cite{wang2008gaussian} or (Gaussian) kernel density estimation \cite{piccardi2007hidden,henter2016kernel}, although in these models the amount of computation per datum scales linearly with the number of training examples, which is infeasible for large datasets.
All of these model classes have been extensively applied to sequence-modelling tasks, but have consistently failed to produce high-quality random samples for complicated data such as motion and speech.
We attribute this shortcoming to the explicit distributional assumptions (e.g., Gaussianity) common to all these models\new{{} -- real data, e.g., motion capture, is seldom Gaussian.}

%All of these paradigms have been extensively used in generative sequence models.
%Unfortunately, these models are still too inflexible to well describe complicated signals such as motion and speech, as can be seen by the poor quality of random samples from these models.
% (cf.\ \cite{uria2015modelling}).
%Deep learning has enabled more advanced autoregressive models of continuous-valued data, such as \cite{graves2013generating,zen2014deep,fragkiadaki2015recurrent,uria2016neural}, where outputs remain explicitly defined as Gaussians or mixture distributions, for tractable inference. These models are however still not sufficiently expressive for many applications.
%Many of the strongest deep and probabilistic autoregressive models currently available, such as \cite{oord2016wavenet,salimans2017pixelcnn,kalchbrenner2018efficient}, model low-dimensional vectors ($\real^3$ or less) in time or space, and it is not clear how they may be scaled up to data such as motion-data sequences with 50 or more dimensions.

Three methods for relaxing the above distributional constraints have gained recent interest.
The first is to quantise the data and then fit a discrete model to it.
Deep autoregressive models on quantised data, such as \citet{oord2016wavenet,salimans2017pixelcnn,kalchbrenner2018efficient,wang2018autoregressive,oord2017neural}, are the state of the art in many low-dimensional ($\real^3$ or less) sequence-modelling problems.
However, it is not clear if these approaches scale up to motion data, with 50 or more dimensions.
Quantisation may also introduce perceptual artefacts.
%Recent research into deep generative models for complex data has also explored two alternative paths:
A second approach is variational autoencoders \cite{kingma2014auto,rezende2014stochastic}, which optimise a variational lower bound on model likelihood while simultaneously learning to perform approximate inference.
The gap between the true maximum likelihood and that achieved by VAEs has been found to be significant \cite{cremer2018inference}.
%The approximations inherent in VAEs create a notable gap from VAEs up to the true maximum likelihood \cite{cremer2018inference}.

The third approach is \new{GANs \cite{goodfellow2014generative,goodfellow2016nips}, that generate samples from complicated distributions \emph{implicitly}, by passing simple random noise through a nonlinear neural network.
As GAN architectures do not allow inference, they are instead trained via} a game against an adversary.
%The third approach is generative adversarial networks \cite{goodfellow2014generative,goodfellow2016nips}, which describe implicit distributions that are easy to sample from but do not allow inference, and instead are trained by means of a game against an adversary.
GANs have produced some very impressive results in image generation \cite{brock2019large}, illustrating the power of implicit sample generation, but their optimisation is fraught with difficulty \cite{mescheder2018training,lucic2018gans}.
GAN output quality usually improves by artificially reducing the generator entropy during sampling, compared to sampling from the distribution actually learned from the data, cf.\ \citet{brock2019large}.
This is often referred to as ``reducing the temperature''.
%The best GAN results increased with artificial means reducing the entropy of the, meaning that samples are not drawn from the distribution.

While VAEs in principle have a partially-implicit generator structure, an issue dubbed ``posterior collapse'' means that VAEs with \emph{strong decoders}, that can represent highly flexible distributions given the latent variable, tend to learn models where latent variables have little impact on the output distribution \cite{chen2017variational,huszar2017maximum,rubenstein2019variational}.
This largely nullifies the benefits of the implicit parts of the generator, leading to blurry and noisy output.

This article considers a less explored methodology called normalising flows \cite{deco1994higher,dinh2015nice,dinh2017density,huang2018neural} \new{(no relation to optical flow)}, especially a variant called Glow \cite{kingma2018glow}, which, \vspace{-0.001pt}like GANs and quantisation, gained attention for highly realistic-looking image samples.
We believe normalising flows offer the best of both worlds, combining a basis in likelihood maximisation and efficient inference like VAEs
%(but without requiring approximations)
with purely implicit generator structures like GANs.
%A recent improvement on normalising flows called Glow \cite{kingma2018glow} grabbed attention by producing perhaps the most realistic-looking image samples thus far from a model trained using maximum likelihood.
Consequently, our paper presents one of the first Glow-based sequence models, and the first to our knowledge to combine autoregression and control, as well as to integrate long memory via a hidden state.
The most closely-related methods are WaveGlow \cite{prenger2019waveglow} and FloWaveNet \cite{kim2019flowavenet} for audio waveforms and VideoFlow \cite{kumar2020videoflow} for video.
We extend these in several novel directions:
Unlike \citet{prenger2019waveglow,kim2019flowavenet}, our architecture is autoregressive \new{(``closed-loop'')}, avoiding costly dilated convolutions \new{and continuity issues (e.g., blocking artefacts) common in open-loop systems, cf.\ \citet{juvela2019gelp}.}
Unlike \citet{kumar2020videoflow}, our architecture permits output control.
In contrast to all three models, we add a recurrent hidden state to enable long memory, which significantly improves the model.
%(The absence of recurrence was hypothesised to cause occlusion-related forgetting in \cite{kumar2020videoflow}).
We also consider data dropout to increase adherence to the control signal.
%We also present a dropout scheme that enhances the consistency of the motion control and the realism of long motion-sequence samples.

%Sequence-modelling applications of these methods are only just emerging, e.g., \cite{prenger2019waveglow,kim2019flowavenet,kumar2020videoflow}.
%, as well as to improve control precision via data dropout.

%A number of hypotheses have been formulated for the success of GANs in generating lifelike output. . We hypothesise that the most important element is the structure of the generator that can describe a highly-varied family of distributions with millions of degrees of freedom. Normalising flows have similarly general .

%\vspace{-0.21ex}
\subsection{Deterministic data-driven motion synthesis}
\label{ssec:motionmodels}
While traditional motion synthesis uses concatenative approaches such as \emph{motion graphs} \cite{arikan2002interactive,kovar2002motion,kovar2004automated}, there has been a strong trend towards statistical approaches.
%act by concatenating short segments from the training database into novel configurations. %While such methods allow control and can be used with relatively small data sets, they do not generalise well and can only produce motion already present in the data. They also suffer from increased latency, as synthesis needs to await a concatenation point before responding to the control signal.
%The work in \cite{sadoughi2019speech} represents a hybrid approach, where data are used to learn a probabilistic model of when to generate motions from a fixed, hand-designed library.
These can roughly be categorised into deterministic and probabilistic methods.
Deterministic methods yield a single prediction for a given scenario, whereas probabilistic methods attempt to describe a range of possible motions.
Deterministically predicted pose sequences usually quickly regress towards the mean pose, cf.\ \citet{fragkiadaki2015recurrent,ferstl2019multi}, since that is the a-priori (i.e., no-information) minimiser of the MSE.
Such methods thus require additional information to disambiguate pose predictions.
Sometimes adding an external control signal suffices -- lip motion is for example highly predictable from speech and has been successfully modelled with deterministic methods \cite{taylor2017deep,suwajanakorn2017synthesizing,karras2017audio}.
Locomotion generation represents a more challenging task, where path-based motion control does not suffice to unambiguously define the overall motion, and simple MSE minimisation results in characters that ``float'' along the control path.
Proposals to overcome this issue in deterministic models include learning and predicting foot contacts \cite{holden2016deep}, or the phase \cite{holden2017phase} or pace \cite{pavllo2018quaternet} of the gait cycle.
\new{\citet{starke2020local} generalised the idea of motion phase to complex motion by letting each bone in a character follow a separate motion phase.}
Autoregressively feeding in previously-generated poses might help combat regression to the mean, and has been used in motion generation without control inputs \cite{fragkiadaki2015recurrent,butepage2017deep,zhou2018auto}.
\citet{zhang2018mode} use a similar approach to generate controllable quadruped motion, letting autoregressive and control information modify network weights, and demonstrate successful generation of both cyclic motion (gait) and simple non-cyclic motion such as jumping.

For many types of motion, no information is readily available that successfully disambiguates motion predictions.
One example is co-speech gestures like head and hand motion, where the motion is unstructured and aperiodic and the dependence on the control signal (speech acoustics or transcriptions) is weak and nonlinear.
The absence of strongly predictive input information means that deterministic motion-generation methods such as \citet{ding2015blstm,hasegawa2018evaluation,kucherenko2019analyzing,yoon2019robots} largely fail to produce distinct and lifelike motion.

%Some methods simply predict or generate motion from previous poses, while others allow the generated motion to be controlled by a separate input signal.

\subsection{Probabilistic data-driven motion synthesis}
\label{ssec:probmotion}
Probabilistic models represent another path to avoid collapsing on a mean pose: By building models of all plausible pose sequences given the available information (prior poses and/or control inputs), any randomly-sampled output sequence should represent convincing motion.
%Control is achieved by conditioning the distribution on other random variables that constitute inputs to the model.
As discussed in Sec.\ \ref{ssec:seqmodels}, many older models assume a Gaussian or Gaussian mixture distribution for poses given the state of the process, for example the (hidden) LSTM state.
Conditional restricted Boltzmann machines (cRBMs) \cite{taylor2009factored,taylor2011two} are one example of this.
The hidden state can also be made probabilistic.
Examples include the SHMMs used for motion generation in \citet{brand2000style}, locally linear models like switching linear dynamic systems (SLDSs) \cite{bregler1997learning,murphy1998switching}, Gaussian processes latent-variable models (GP-LVMs) \cite{lawrence2005probabilistic}, and VAEs \cite{kingma2014auto,rezende2014stochastic}.
Locally linear models were used for for motion synthesis in \citet{pavlovic2000learning,chai2005performance}, but have primarily been applied in recognition tasks.
GP-LVMs and the closely related Gaussian process dynamical models (GPDMs) have been extensively studied in motion generation \cite{grochow2004style,wang2008gaussian,levine2012continuous} but they -- along with other kernel-based motion-generation methods such as the radial basis functions (RBFs) in \citet{rose1998verbs,kovar2004automated,mukai2005geostatistical} -- are unattractive in the big-data era since their memory and computation demands scale quadratically (or worse) in the number of training examples.
%in a na\"{\i}ve implementation.
%One classic branch of probabilistic machine-learning for motion is Gaussian processes latent-variable models (GP-LVMs) \cite{lawrence2005probabilistic}.
%Many of these models (but not all \cite{grochow2004style}) incorporate a hidden state across time, e.g., \cite{wang2008gaussian,levine2012continuous}.
VAEs circumvent computational issues by using a variational and amortised (see \citet{cremer2018inference}) approximation of the likelihood for training.
They have been applied to model \new{controllable} human locomotion \new{\cite{habibie2017recurrent,ling2020character}} and to generate head motion from speech \cite{greenwood2017predicting,greenwood2017predictingb}.
\new{\citet{ling2020character} describes an autogregressive unconditional motion model based on VAEs, using a deterministic decoder based on the mixture-of-experts architecture from \citet{zhang2018mode}.
$\beta$-VAEs \cite{higgins2016beta} are used to mitigate posterior collapse, while scheduled sampling \cite{bengio2015scheduled} is necessary to stabilise long-term motion generation.
Reinforcement learning is used to enable character control, although response time is somewhat sluggish.}
Notably, many VAE methods either generate noisy motion samples (e.g., \citet{taylor2011two}) or choose to not sample from the (Gaussian) observation distribution given the latent state of the process,
%or avoid this by not sampling from the Gaussian next-step distribution at every time frame given the deterministic or stochastic process state,
instead generating the mean of the conditional Gaussian only \new{\cite{greenwood2017predicting,greenwood2017predictingb,ling2020character}}.
This risks re-introducing mean collapse and artificially reduces output entropy.
We take this as evidence that these methods failed to learn an accurate and convincing motion distribution.

%As mentioned in Sec.\ \ref{ssec:seqmodels}, GAN-based approaches such as \cite{brock2019large} are considered the state of the art in complex data-generation tasks such as image generation.
Variations of GANs \cite{sadoughi2018novel} and adversarial training \new{\cite{wang2019combining,ferstl2019multi,starke2020local}} have also been applied for motion generation and the related task of generating speech-driven video of talking faces \cite{vougioukas2018end,pumarola2018ganimation,pham2018generative,vougioukas2020realistic}.
\new{In contrast to GANs and VAEs, \citet{starke2020local} add latent-space noise to motion only at synthesis time (not during training), to obtain more varied motion, albeit at the expense of deviating from the desired input control.
This approach also means that the distribution of the motion is not learned, and need not match that of natural motion.}

Unlike previously-cited probabilistic motion-generation methods, GANs do not assume that observations are Gaussian given the state of the data-generating process.
This avoids both regression towards the mean and Gaussian noise in output samples.
The same goes for the discretisation-based approach in \citet{sadoughi2019speech}, which learns a probabilistic model
%of when to trigger the playback of sequences
that triggers motion sequences from a fixed
%, hand-designed
motion library.
%Normalising flows share this advantage with GANs, but can be trained using conventional maximum likelihood, similar to most probabilistic models cited above.
We consider another method for avoiding Gaussian assumptions, by introducing the first probabilistic motion model based on normalising flows.
%This provides advantages in terms of control, responsiveness, stability, and generality.
\new{%
%In contrast to \citet{starke2020local}, our model is not just stochastic but probabilistic, and has no compromise between variety and control accuracy; in contrast to \citet{ling2020character}, control is built into our model.
In contrast to MVAEs \cite{ling2020character}, our method can model conditional motion distributions, and so has controllability built in.}

\section{Method}
\label{sec:method}
%In this section we describe the mathematical basis of autoregressive models and normalising flows (particularly Glow), and how we adapt them to generate controllable motion data.
%In this section we review the mathematical basis of normalising flows, particularly Glow, and show how we adapt them to generate controllable motion data.
\newer{This section introduces our new probabilistic motion model.
The basic idea is to treat motion as a series of poses, and model these poses using an autoregressive model.
In other words, we describe the conditional probability distribution of the next pose in the sequence as a function of previous poses and relevant control inputs.
%MoGlow is an autoregressive sequence model, that describes the conditional probability distribution of the next pose as a function of previous poses and relevant control inputs.
Like in a conditional GAN, the next pose of the motion is generated by drawing a random sample from a simple distribution such as a Gaussian, and then nonlinearly transforming that sample by passing it through a neural network.
This has the effect of reshaping the simple starting distribution into a more complex distribution that fits the distribution of the next pose in data.
However, unlike a GAN, the neural network we use is invertible, which allows us to directly compute and maximise the likelihood of the data under the model.
This makes the model stable to train.
We now introduce basic notation and (in Sec.\ \ref{ssec:flows}) describe how to construct normalising flows.
Secs.\ \ref{sec:moglow} and \ref{ssec:dropout} then detail, step by step, how to build a controllable autoregressive sequence model out of such flows.}
\begin{figure}
\centering
\includegraphics[width=0.825\columnwidth]{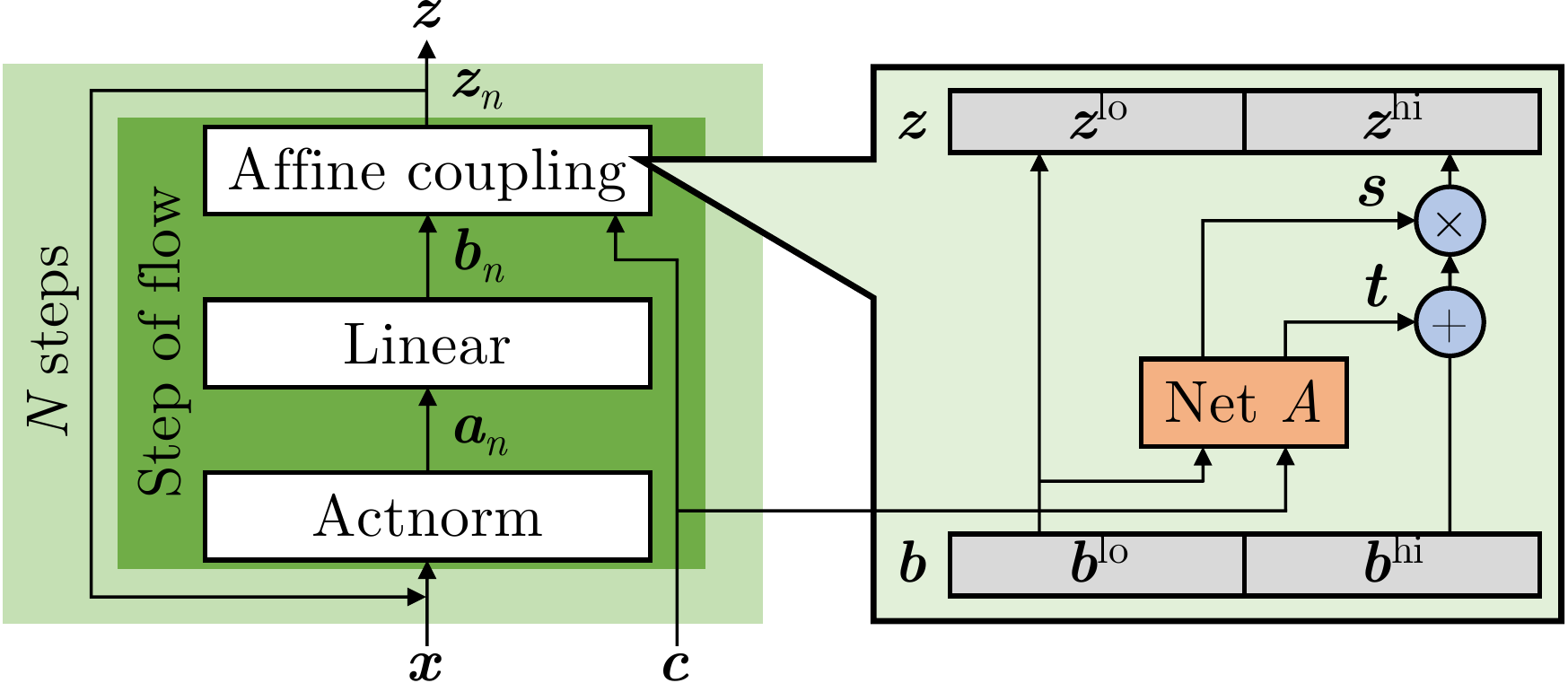}
\vspace{-1.2ex}
\caption{Glow steps $\f^{-1}_n$ during inference. Detail of coupling layer on right.}
\label{fig:glow}
\vspace{-2.3ex}
\Description{A block diagram showing how the three sub-steps actnorm, linear transformation, and affine coupling are repeated N times (steps) when transforming x to z. An inset contains a block diagram of the computations inside the affine coupling layer, showing how input elements with low indices are passed straight through without transformation, but also (together with the conditioning information) fed into a neural network A. The neural network outputs vectors of additive offsets t and scaling factors s that are used to affinely transform the input elements to the coupling with high indices.}
\end{figure}

%\subsection{Preliminary notation}
%\label{ssec:notation}
For notation, we write vectors, and sequences thereof, in bold font.
Upper case is used for random variables and matrices, and lower case for deterministic quantities or specific outcomes of the random variables.
In particular, $\X$ typically represents randomly-distributed motion with $\x\in\real^{D\times{}T}$ being an outcome of the same\newer{, while $\ctrl\in\real^{C\times{}T}$ represents the matching \emph{control-signal inputs}, which in our experiments are relative and rotational velocities that describe motion along path on the ground plane.}
Non-bold capital letters generally denote indexing ranges, with matching lower-case letters representing the indices themselves, e.g., $t\in\{1,\,\ldots,\,T\}$.
Indices into sequences extract specific time frames, for example individual \emph{poses} $\x_t\in\real^D$, or sub-sequences $\xrange{1}{t}=[\x_1,\,\ldots,\,\x_t]$.
\newer{Each pose parameterises the positions and orientations of objects such as a whole body, parts of a body, or keypoints on a body or face.
In this paper, the pose vector $\x_t$ is created by concatenating vectors that represent either joint positions or joint rotations on a 3D skeleton.}
%\emph{Motion} $\x$ is represented as a time series of \emph{poses} $\x_t$ that parameterise the 2D or 3D positions and orientations of objects such as a whole body, parts of a body, or keypoints on a body or face.

%\subsection{Temporal structure of the model}

%\newer{Fundamentally, MoGlow is an autoregressive model, that describes $\X_t$, the conditional probability distribution of the next pose, as a function of previous poses $\x_{1:t-1}$ and of relevant control inputs $\ctrl$ (here $\ctrl_{1:t}$ to avoid algorithmic latency).
%Like in a conditional GAN, the next pose in the sequence is generated by drawing a random sample $\z_t$ from a simple distribution such as a Gaussian, and then nonlinearly transforming that vector by passing it through a neural network, that additionally takes conditioning information as input.
%This has the effect of reshaping the distribution $\Z_t$ into a distribution that fits the distribution of the next pose in data.
%However, unlike a GAN, the neural network we use is invertible, which allows us to directly compute and maximise the likelihood of the data under the model, and makes the model stable to train.
%We now (Sec.\ \ref{ssec:flows}) describe how to construct normalising flows, and then describe step by step how to construct a controllable autoregressive sequence model out of such flows (Secs.\ \ref{sec:moglow} and \ref{ssec:dropout}).}

\subsection{Normalising flows and Glow}
\label{ssec:flows}
Normalising flows are flexible generative models that allow both efficient sampling and efficient inference. %(likelihood computation). 
The idea is to subject samples from a simple, fixed \emph{base} (or \emph{latent}) distribution $\Z$ on $\real^D$ to an invertible and differentiable nonlinear transformation $\f:\real^D\to\real^D$, in order to produce samples from a new, more complex distribution $\X$. 
%The transformation $\f$ is parameterised by some $\bb{\theta}$. 
If this transformation has many degrees of freedom, a wide variety of different distributions can be described.

%Like in deep learning in general, 
Flows construct expressive transformations $\f$ by chaining together numerous simpler nonlinear transformations $\{\f_n\}_{n=1}^N$, each of them parameterised by a $\bb{\theta}_n$ such that $\bb{\theta}=\{\bb{\theta}_n\}_{n=1}^N$.
We define the observable random variable $\X$, the latent random variable $\Z\sim \Gauss\left(\bb{0},\,\bb{I}\right)$, and intermediate distributions $\Z_n$ as follows:
\begin{align}
\z & = \z_N \overset{\f_N}{\to} \z_{N-1} \overset{\f_{N-1}}{\to} \ldots \overset{\f_2}{\to} \z_1 \overset{\f_1}{\to} \z_0 = \x\\
\x & = \f(\z) = \f_1 \left( \f_2 \left( \ldots \f_N\left(\z\right) \right) \right)\\
%\x & = \f(\z) = \f_1 \circ \f_2 \circ \ldots \circ \f_N(\z)\\
\z_n\left(\x\right) & = \f_n^{-1} \circ \ldots \circ \f_1^{-1}(\x)
\label{eq:flowvars}
\text{.}
\end{align}
The sequence of (inverse) transformations $\f^{-1}_n$ in \eqref{eq:flowvars} is known as a \emph{normalising flow}, since it transforms the distribution $\X$ into an isotropic standard normal random variable $\Z$.
%Since the mapping $\f$ is composed from a sequence of simple nonlinear mappings, it is sometimes known as an \emph{invertible neural network}.

Similar to the generators in GANs, normalising flows are \emph{implicit probabilistic models} according to the definition in \citet{mohamed2016learning}.
While explicit models draw samples from probability density functions defined in the space of the observations, GANs and normalising flows instead generate output by drawing samples $\z$ from a latent base distribution $\Z$ that acts as a source of entropy, and then subjecting these samples to a deterministic, nonlinear transformation $\f$ to obtain samples $\x=\f(\z)$ from $\X$.
%defined not by a probability density function in the space of the observations $\X$ but as a nonlinear transformation $\f$ of a latent distribution $\Z$.
%Generators of this form.
Unlike GANs, however, normalising flows permit
%tractable and efficient
\new{fast and easy probability computation (inference)}, since the transformation $\f$ is invertible:
Using the change-of-variables formula, we can write the log-likelihood of a sample $\x$\new{, as used in likelihood maximisation,} as
\begin{align}
\ln p_{\bb{\theta}}\left(\x\right) & = \ln p_{\Gauss}\left(\z_N\left(\x\right)\right) + \sum_{n=1}^N \ln \left\vert \det \frac{\partial \z_n\left(\x\right)}{\partial \z_{n-1}} \right\vert
\label{eq:flowlnp}
\text{,}
\end{align}
where $\frac{\partial \z_n\left(\x\right)}{\partial \z_{n-1}}$ is the Jacobian matrix of $\f^{-1}_n$ at $\x$, which depends on $\bb{\theta}$, and $p_{\Gauss}$ is the probability density function of the $D$-dimensional standard normal distribution.
The general determinant in \eqref{eq:flowlnp} has computational complexity close to $\mathcal{O}(D^3)$, so many improvements to normalising flows involve the development of $\f_n$-transformations with tractable Jacobian determinants, that nonetheless yield highly flexible transformations under iterated composition.
An
%accessible and
in-depth review of normalising flows and different flow architectures can be found in \citet{papamakarios2019normalizing}.
%One example of this is \emph{Glow} \cite{kingma2018glow}.
In this work, we consider the \emph{Glow} architecture \cite{kingma2018glow}, first developed for images, and extend it to model controllable motion sequences.

%\subsection{Glow}
%\label{ssec:glow}
Each component transformation $\f^{-1}_n$ in Glow contains three sub-steps: \emph{activation normalisation}, also known as \emph{actnorm}; a \emph{linear transformation}; and a so-called \emph{affine coupling layer}, together shown as a \emph{step} of flow in in Fig.\ \ref{fig:glow}.
The first two are affine or linear transformations while the latter amounts to a more powerful nonlinear transformation that is nonetheless invertible.

We will let $\bb{a}_{t,\,n}$ and $\bb{b}_{t,\,n}$ denote intermediate results of Glow computations for observation $\x_t$ in flow step $n$, as shown in Fig.\ \ref{fig:glow}.
Actnorm, the first sub-step, is an affine transformation $\bb{a}_{t,\,n}=\bb{s}_n\odot\z_{t,\,n-1}+\bb{t}_n$ (with $\odot$ denoting elementwise multiplication) intended as a substitute for batchnorm \cite{ioffe2015batch}.
%Reminiscent of FixUp \cite{zhang2019fixup},
The parameters $\bb{s}_n>0$ and $\bb{t}_n$ are initialised such that the output has zero mean and unit variance and then treated as trainable parameters.
% (elements of $\bb{\theta}_n$).
After actnorm follows a linear transformation $\bb{b}_{t,\,n}=\bb{W}_n\bb{a}_{t,\,n}$ where $\bb{W}\in\real^{D\times{}D}$.
%For Glow on images, this matrix measures $c\times{}c$ and is applied individually to a $W\times{}H$ grid of $c$-vectors.
%For our time-series data, we apply the matrix $\bb{W}_n$ independently to the vectors $\bb{a}_{t,\,n}$ for each $t$.
%For images, this is convolved over the $x$ and $y$ directions, but we will only apply it along the time dimension $t$.
%This is applied to isolated groups of $D$ variables through a convolution with filter-size one.
By representing $\bb{W}_n$ by an LU-decomposition $\bb{W}_n=\bb{L}_n\bb{U}_n$ with one matrix diagonal set to one (say $l_{n,\,dd}=1$), the Jacobian determinant of the sub-step is just the product of the diagonal elements $u_{n,\,dd}$, which is computable in linear time.
The non-fixed elements of $\bb{L}_n$ and $\bb{U}_n$ are the trainable parameters of the sub-step.% and elements of $\bb{\theta}_n$.

The affine coupling layer is more complex.
The idea is to affinely transform half of the input elements based on the values of the other half.
By passing those remaining elements through unchanged, it is easy to use their values to undo the transformation when reversing the computation.
Mathematically, we define $\bb{b}_{t,\,n}$ and $\z_{t,\,n}$ as concatenations $\bb{b}_{t,\,n}=[\bb{b}_{t,\,n}^{\mathrm{lo}},\,\bb{b}_{t,\,n}^{\mathrm{hi}}]$ and $\z_{t,\,n}=[\z_{t,\,n}^{\mathrm{lo}},\,\z_{t,\,n}^{\mathrm{hi}}]$.
The coupling can then be written
\begin{align}
[\z_{t,\,n}^{\mathrm{lo}},\,\z_{t,\,n}^{\mathrm{hi}}]
& = [\bb{b}_{t,\,n}^{\mathrm{lo}},\,
(\bb{b}_{t,\,n}^{\mathrm{hi}}+\bb{t}_{t,\,n}^\prime)\odot\bb{s}_{t,\,n}^\prime]
\label{eq:coupling}
%\intertext{where}
%[\bb{s}\left(\x_1\right),\,\bb{t}\left(\x_1\right)]
%&= \bb{g}\left(\x_1\right)
\text{.}
\end{align}
The scaling $\bb{s}_n^\prime>\bb{0}$ and bias $\bb{t}_n^\prime$ terms in the affine transformation of the $\bb{b}_{t,\,n}^{\mathrm{hi}}$ are computed via a neural network, $A_n$, that only takes $\bb{b}_{t,\,n}^{\mathrm{lo}}$ as input.
%\begin{align}
%[\bb{s}_{t,\,n}^\prime,\,\bb{t}_{t,\,n}^\prime]
%& = A_n\left(\bb{b}_{t,\,n}^{\mathrm{lo}}\right)
%\label{eq:unconditionalcoupling}
%\text{.}
%\end{align}
(We use `$A$' for ``affine''.)
We can therefore unambiguously invert Eq.\ \eqref{eq:coupling} based on $\z_{t,\,n}$ by feeding $\z_{t,\,n}^{\mathrm{lo}}=\bb{b}_{t,\,n}^{\mathrm{lo}}$ into $A_n$ to compute $\bb{s}_n^\prime>\bb{0}$ and $\bb{t}_n^\prime$.
%Because $\bb{b}_{t,\,n}^{\mathrm{lo}}$ passes through the network unchanged, w
The coupling computations during inference are visualised in Fig.\ \ref{fig:glow}.
%Since $\z_{t,\,n+1}^{\mathrm{lo}}=\bb{b}_{t,\,n}^{\mathrm{lo}}$, the scale and bias terms needed to invert the affine coupling layer are computable unambiguously from $\z_{t,\,n+1}$ also by forward propagation through $A_n$.
The weights that define $A_n$ are also elements of the parameter set $\bb{\theta}_n$, while the constraint $\bb{s}_n^\prime>0$ is enforced by using a sigmoid nonlinearity \cite[App.\ D]{nalisnick2019deep}.
Random weights are used for initialisation except in the output layer, which is initialised to zero \cite{kingma2018glow}; this has the effect that the coupling initially is close to an identity transformation, reminiscent of Fixup initialisation \cite{zhang2019fixup}.

Interleaved linear transformations and couplings are both necessary for an expressive flow.
Without couplings, a stack of flows collapses to compute a single, fixed affine transformation of $\Z$, meaning that $\X$ will be restricted to a Gaussian distribution; a stack of couplings alone will only perform a nonlinear transformation of \emph{half} of $\Z$, doing nothing to the other half.
The linear layers $\bb{W}_n$ can be seen as generalised permutation operations between couplings, ensuring that all variables (not just one half) can be nonlinearly transformed with respect to each other by the full flow.

%\subsection{Integrating Glow into the temporal model}
\subsection{MoGlow}
\label{sec:moglow}
%\label{ssec:armodels}
Let $\X=\Xrange{1}{T}=[\X_1,\,\ldots,\,\X_T]$ be a sequence-valued random variable.
Like all autoregressive models of time sequences, we develop our model from the %(always valid) 
decomposition
\begin{align}
p\left(\x\right)
& = p\left(\xrange{1}{\tau}\right)
\prod_{t=\tau+1}^T p\given{\x_t}{\xrange{1}{t-1}}
\label{eq:decomposition}
\text{.}
\end{align}
%This decomposition is the foundation of autoregressive models of stochastic processes.
%If one assumes that the distribution $\X_t$ only depends on the $\tau$ previous values, so that $p\given{\x_t}{\xrange{1}{t-1}} \equiv p\given{\x_t}{\xrange{t-\tau}{t-1}}$, one obtains a process known as a \emph{Markov chain of order $\tau$}.
%We call $p\given{\x_t}{\xrange{t-\tau}{t-1}}$ the \emph{next-step distribution}.
%Longer memory can be added to Markov models by introducing a \emph{latent} (unobservable) state $\bb{h}_t\in\real^H$ which affects the distribution of the observable $\x_t$ and evolves according to a function $\bb{g}$ at each timestep.
%The result is a (hidden) \emph{state-space model}; cf.\ Sec.\ \ref{ssec:seqmodels}.
We assume the distribution $\X_t$ only depends on the $\tau$ previous values (i.e., is a Markov chain of order $\tau$), except for a latent state $\bb{h}_t\in\real^H$ that \new{represents the effect of recurrence in a recurrent neural network (RNN) and} evolves according to a relation $\bb{h}_{t}=\bb{g}\left(\bb{h}_{t-1}\right)$ at each timestep.
%\new{that represents the effect of recurrence in a recurrent neural network (RNN).}
%Thus, the next-step distribution is given by $p\given{\x_t}{\xrange{t-\tau}{t-1},\bb{h}_t}$.
%In the vast majority of data-generation (synthesis) applications, output should not only be natural, but also must satisfy certain constraints defined by the application.
%In motion synthesis, one might for instance want locomotion to follow a certain path, body language to express a certain emotion, or face and body motion to match a spoken message.
To achieve control over the output we further condition the $\X$-distribution on another sequence variable $\Ctrl$, acting as the \emph{control signal}.
We assume that, for each training-data frame $\x_t$, the matching control-signal values $\ctrl_t\in\real^{C}$ are known.
%(``Global'' control parameters that are constant across an entire output sequence can easily be represented in this framework by including elements in $\ctrl_t$ that are kept fixed across each sequence.)
Moreover, the experiments in this paper focus on causal control schemes, where only current and former control inputs $\crange{1}{t}$ may influence the conditional distributions from \eqref{eq:decomposition} at $t$.
(Letting the model also depend on future $\ctrl$-values might improve motion quality, but inevitably introduces algorithmic latency.)
Putting the Markov assumption, the hidden state, and the control together gives our temporal model
\begin{align}
p_{\bb{\theta}}\given{\x}{\ctrl}
& = p\given{\xrange{1}{\tau}}{\crange{1}{\tau}}
\prod_{t=\tau+1}^T p_{\bb{\theta}}\given{\x_t}{\xrange{t-\tau}{t-1},\,\crange{t-\tau}{t},\,\bb{h}_{t-1}}
\label{eq:moglowdist}\\
\bb{h}_{t} & = \bb{g}_{\bb{\theta}}\left(\xrange{t-\tau}{t-1},\,\crange{t-\tau}{t},\,\bb{h}_{t-1}\right)
\label{eq:moglowstate}%\\
%[\bb{s}_{t,\,n}^\prime,\,\bb{t}_{t,\,n}^\prime]
%& = A_n\left(\bb{b}_{t,\,n}^{\mathrm{lo}},\,\xrange{t-\tau}{t-1},\,\ctrl_{t-\tau:t,\,n},\,\bb{h}_t\right)
%\label{eq:moglowcoupling}
\text{,}
\end{align}
where we have decided to condition on the control signal at most $\tau$ steps back only, just like for the previous poses.
The subscript $p_{\bb{\theta}}$ indicates that the distributions depend on model parameters $\bb{\theta}$.
The initial hidden state can be learned, but in our experiments we initialise $\bb{h}_{\tau}$ as $\bb{0}$.%
\footnote{For this article, we will ignore how to model the initial distribution $p\left(\xrange{1}{\tau}\right)$ from \eqref{eq:moglowdist}. %
%and will exclude its contribution to any probability computations.
Experimentally, we found that initialisation with natural motion snippets or with a static mean pose both give competitive results.}
For the deterministic hidden-state evolution $\bb{g}$ a straightforward choice to implement Eq.\ \eqref{eq:moglowstate} is to use a recurrent neural network, here an LSTM \cite{hochreiter1997long}.
\new{The vector $\bb{h}_t$ is then the concatenation of the LSTM cell state vectors and the LSTM-unit output vectors at time $t$.}

Finally, we also assume \emph{stationarity}, meaning that $\bb{g}$ and the distributions in \eqref{eq:moglowdist} are independent of $t$.
This is an exceedingly common assumption in practical sequence models, since it means that all timesteps in the training data can be treated as samples from a single, time-independent distribution $p_{\bb{\theta}}\given{\x_t}{\xrange{t-\tau}{t-1},\,\crange{t-\tau}{t},\,\bb{h}_{t-1}}$.
The central innovation in this paper is to learn that controllable \emph{next-step distribution} using normalising flows.
%For many processes, including motion, it is reasonable to assume that the laws that govern the process are the same at any given point in time, meaning that the next step-distribution and $\bb{g}$ are independent of $t$.
%Mathematically, $p\given{\x}{\xrange{t-\tau}{t-1}} \equiv p\given{\x}{\xrange{t'-\tau}{t'-1}}$, whenever
%This assumption is known as \emph{stationarity} and 

%This model has complete memory $\tau$ steps back in time, but can also model dependencies further back in time (theoretically all the way back to $\x_1$) thanks to the hidden state variable.
%So-called hidden Markov models (HMMs) and Kalman filters are special cases of such models in which $\bb{g}$ is a stochastic function.

%For many processes, including motion, it is reasonable to assume that the laws that govern the process are the same at any given point in time, meaning that the next step-distribution is independent of $t$.

To adapt Glow to parameterise the next-step distribution in the autoregressive hidden-state model in Eqs.\ \eqref{eq:moglowdist} and \eqref{eq:moglowstate}, we made a number of changes to the original image-oriented Glow architecture in \citet{kingma2018glow}.
There, dependencies between $\Z_{t,\,n}$-values at different image locations were introduced by making $A_n$ a convolutional neural network.
We instead use unidirectional (causal) LSTMs inside $A_n$ to enable dependence between timesteps, which is simpler than the dilated convolutions used in recent audio models based on Glow \cite{prenger2019waveglow,kim2019flowavenet} while giving better models than making $A_n$ a simple feedforward network.
%This change is consistent with recent developments in generative modelling of audio waveforms, which is seeing a shift away from dilated convolutions as introduced in WaveNet \cite{oord2016wavenet} towards more computationally attractive recurrent approaches such as WaveRNN \cite{kalchbrenner2018efficient} and its relatives \cite{valin2019lpcnet,lorenzo2019towards}.

We added a small epsilon $\varepsilon=0.05$ to the sigmoids in $A_n$ that define the scale-factor outputs $\bb{s}_n^\prime$, in order to bound the dynamic range of the scaling and stabilise training.
This modification restricts the possible scale-factor values to the interval $\bb{s}_n\in(\varepsilon,1+\varepsilon)$.
Unlike \citet{dinh2017density,kim2019flowavenet}
%,razavi2019generating}
we did not use any multiresolution architecture in our flow, as that did not provide any noticeable improvements in preliminary experiments, nor do we include squeeze operations, as that would add algorithmic latency.

%Unlike the similar VideoFlow \cite{kumar2020videoflow} model we also consider external control of the generated sequences.
%Unlike the control scheme is \cite{kingma2018glow}, we jointly learn the distribution and how to control it.
To provide motion control and enable explicit dependence on recent pose history in Glow distributions, we take inspiration from recent sequence-to-sequence audio models \citet{prenger2019waveglow,kim2019flowavenet}, which feed the conditioning information (here $\xrange{t-\tau}{t-1}$ and $\crange{t-\tau}{t}$) as additional inputs to the affine couplings $A_n$,
%\footnote{This is conceptually equivalent to conditioning on $\zrange{t-\tau}{t-1}$ instead of $\xrange{t-\tau}{t-1}$, as the mapping $\x_t=\f\left(\z_t\right)$ is invertible.}
these being the only neural networks in Glow.
The scaling and bias terms\new{, together with the next state ${h}_{t,\,n}$ of net $A_n$,} are then computed as
\begin{align}
[\bb{s}_{t,\,n}^\prime,\,\bb{t}_{t,\,n}^\prime,\,\bb{h}_{t,\,n}]
& = A_n(\bb{b}_{t,\,n}^{\mathrm{lo}},\,\xrange{t-\tau}{t-1},\,\crange{t-\tau}{t},\,\bb{h}_{t-1,\,n})
\label{eq:moglowcoupling}
\text{.}
\end{align}
%(In addition, the output of $A_n$ also depends on the LSTM state $\bb{h}_{t-1,\,n}$.)
We call our proposed model structure \mbox{\emph{MoGlow}} for \emph{motion Glow}.

\begin{figure}
\centering
\includegraphics[width=0.95\columnwidth]{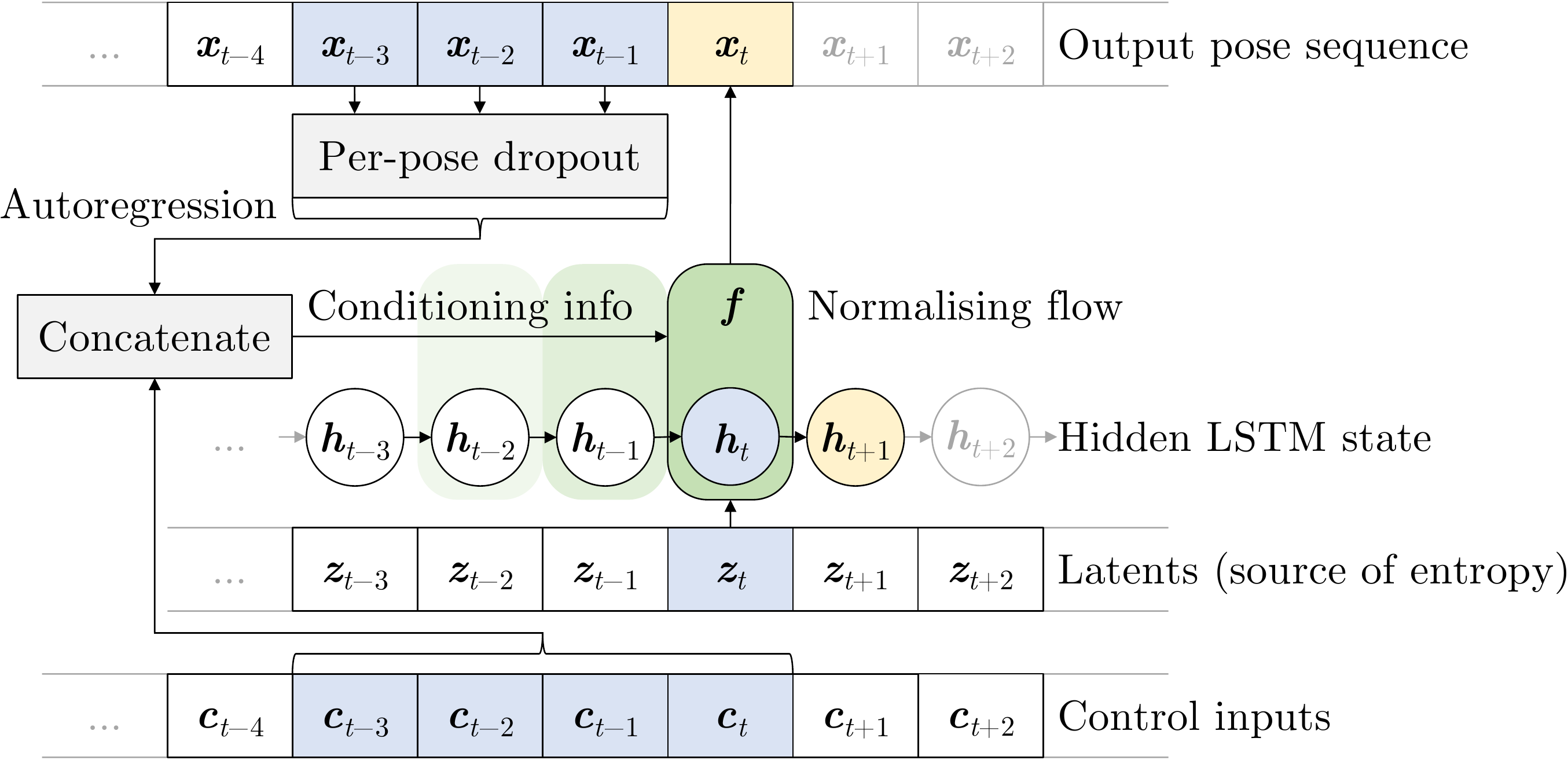}
\vspace{-1.2ex}
\caption{Schematic of autoregressive motion generation with MoGlow. \new{Inputs are blue, outputs yellow.} Dropout is only applied at training time.}
\label{fig:moglow}
\vspace{-2.3ex}
\Description{A block diagram of MoGlow showing how conditioning information, namely recent poses (with per-frame dropout) and recent and current control-input values, along with the hidden state of the RNN can influence how the normalising flow transforms the driving noise z into the sampled pose for the current time frame.}
\end{figure}

If we let $\z_{t,\,N}$ denote the observation $\x_t$ mapped back onto the latent space by the (conditional) flow transformation $\f^{-1}$, the full log-likelihood \new{training objective} of MoGlow applied to a sequence $\x$ given the control input $\ctrl$ can be written
\begin{multline}
\ln p_{\bb{\theta}}\given{\xrange{\tau+1}{T}}{\xrange{1}{\tau},\ctrl}
= \sum_{t=\tau+1}^{T}\ln p_{\Gauss}\left(\z_{t,\,N}\left(\xrange{1}{t},\crange{1}{t}\right)\right)\\
+ \sum_{n=1}^N \sum_{d=1}^{D} \sum_{t=\tau+1}^{T} \left(\ln s_{n,\,d} + \ln u_{n,\,dd} + \ln s_{t,\,n,\,d}^\prime\left(\xrange{1}{t},\,\crange{1}{t}\right) \right)
\label{eq:glowlnp}
\text{,}
\end{multline}
where we have made explicit which terms depend on $\x$ and $\ctrl$.
%It is straightforward to maximise this likelihood in modern machine-learning frameworks like TensorFlow or Torch.
A schematic illustration of MoGlow sample generation is presented in Fig.\ \ref{fig:moglow}.
At generation time, latent $\z_t$-vectors are sampled independently from $p_{\Gauss}$ (acting as a source of randomness for the next-step distribution) and then transformed into new poses $\x_t$ by the flow $\f$ conditioned on $\xrange{t-\tau}{t-1}$, $\crange{t-\tau}{t-1}$, and $\bb{h}_{t-1}$.

\new{Because $\Z_t$ is supported on all of $\real^D$, so is $\X_t$.
This is a natural fit for pose representations that take values on $\real^D$, e.g., joint positions in Cartesian coordinates.
Pose representations supported on a non-zero volume subset $\mathcal{X}\subset\real^D$, for example the exponential map \cite{grassia1998practical}, can also be used.
In practise, we recommend parameterisations that minimise angular discontinuities\newer{, e.g., by expressing angles relative to a T-pose and wrapping at $\pm$180 degrees}, since the method works best for continuous density functions.
%Since the range of real joint angles is limited, away from discontinuities.
%The architecture is not directly applicable to zero-volume manifolds.
%, such as joint positions on a sphere of perfectly constant bone length.
}

\subsection{Data dropout}
\label{ssec:dropout}
Early MoGlow models had a problem with poor adherence to the control input,
%, compared to the autoregressive input.
where generated character motion often would walk or run even when the control input (in this case, the path followed by the root node) specified that no movement through space was taking place.
This indicates an over-reliance on autoregressive pose information, compared to the control input.
Such behaviour is a frequent issue with long-term prediction in powerful autoregressive models (cf.\ \citet{chen2017variational,liu2019maximizing}),
%A practical issue with optimising autoregressive models of slowly-changing sequences is that the optimisation might not successfully integrate all the pertinent information from the input, for instance getting stuck on predicting $\x_t=\x_{t-1}$, while ignoring other inputs such as the control signal.
%This is a common stumbling block
for example in generative models of speech as in \citet{uria2015modelling,wang2018autoregressive,tachibana2018efficiently}.
Established methods to counter this failure mode include applying dropout to entire frames of autoregressive history inputs -- conventionally called \emph{data dropout} -- as in \citet{bowman2016generating,wang2018autoregressive}, or downsampling the data sequences as in \citet{tachibana2018efficiently}.
%The squeeze operations in \cite{prenger2019waveglow,kim2019flowavenet} similarly amount to a kind of downsampling.
Dropout and bottlenecks in the autoregressive path can also be combined with a lowered frame rate, e.g., \citet{wang2017tacotron,shen2018natural}.
All of these approaches have the net effect of reducing the informational value of the most-recent autoregressive feedback, thus making the information in the current control input relatively more valuable.
%The squeeze operations in \cite{dinh2017density,kingma2018glow,prenger2019waveglow,kim2019flowavenet} can be seen as another way to reduce the information shared between adjacent outputs, although only \cite{prenger2019waveglow,kim2019flowavenet} apply these squeezes in an autoregressive model.
We found that applying data dropout during training substantially improved the consistency between the generated motion and the control signal in MoGlow models.
%Without dropout, generated motion often walked or ran even when the control signal indicated that no movement through space was taking place.
In particular, the issue of MoGlow running in place vanished with frame dropout rates of 50\% and above.
%Dropout was also found to remedy issues with exposure bias %\cite{ranzato2016sequence}:
%, which is a thorny issue with models that are trained on inputs and outputs derived from ground-truth sequences (\emph{teacher forcing}) but evaluated on long sequences of iterated model samples/predictions \cite{ranzato2016sequence}.
%While early autoregressive MoGlow models tended to revert towards a static pose when sampling sequences several seconds in duration, this issue virtually disappeared after applying dropout to the autoregressive history inputs.
%For the MG configuration we set the frame dropout rate to 0.95.
%It was not necessary to look into more complex solutions such as professor forcing \cite{lamb2016professor}.

\section{Experimental setup}
\label{sec:experiment}
%This section describes the data and systems used in our experiments.
%Evaluations, their results, and their interpretation follow in Sec.\ \ref{sec:results}.

%This section describes an application -- and subsequent evaluation -- of the methods from Secs.\ \ref{sec:method} and \ref{sec:moglow} on a motion-capture dataset of human locomotion.
%We stress that, unlike, e.g., \cite{holden2017phase}, there is nothing in our approach that requires the motion to be (quasi-)periodic like human gait.

%The $D$ dimensions commonly encode each pose as joint rotations or Euclidean coordinates of joints or other keypoints (like in the experiments in Sec.\ \ref{sec:experiment}), although movement in space requires also specifying the position and orientation of the root node(s) in the hierarchy.

The goal of MoGlow is to introduce a probabilistic and controllable motion model capable of delivering high-quality output without task-specific assumptions.
\new{This section presents data and systems used for comparative experiments that evaluate the quality of MoGlow output across different tasks.
Associated evaluations and results are reported in Sec.\ \ref{sec:results}, along with skinned-character experiments designed to validate the probabilistic aspects of the model.
%and consider applications to non-locomotion scenarios.
}
%The aim of our empirical evaluation, described in this section, is thus to demonstrate the 1) general applicability and 2) high quality of the method, particularly with respect to other task-agnostic motion-generation approaches.
%We evaluate the proposed model against other approaches both objectively and in a user study.
%\new{This is complemented by experiments validating the probabilistic aspects of the model and exploring applications beyond locomotion.}

Objectively evaluating motion plausibility is difficult in the general case, as there is no single natural realisation of the motion given typical, weak control signals.
Comparing low-level properties such as frame-wise joint positions between recorded and synthesised motion is therefore not particularly informative.
To enable meaningful objective evaluation, we chose to evaluate MoGlow on locomotion synthesis, for which some perceptually-salient aspects of the motion can be studied objectively.
Specifically, foot-ground contacts are easy to identify as they should have zero velocity, and foot-sliding artefacts (often attributable to mean collapse) are both pervasive in synthetic locomotion and known to greatly affect the perceived naturalness of the resulting animation.
We stress that unlike \new{\citet{holden2016deep,holden2017phase,pavllo2018quaternet,starke2020local}}, we do not use foot-contact information as part of our model, but only use it to objectively evaluate the generated output motion.

\subsection{Data for objective and subjective evaluations}
\label{ssec:data}
We considered two sources of motion-capture data in our evaluations, namely human (bipedal) and animal (quadrupedal) locomotion on flat surfaces.
Bipedal and quadrupedal locomotion represent significantly different modelling problems, and to our knowledge no method has been demonstrated to perform well on both tasks\new{, with the exception of \citet{starke2020local}, which appeared while this paper was in review.}
For the human data, we used the data and preprocessing code provided by \citet{holden2015learning,holden2016deep}.%
\footnote{Please see \href{http://theorangeduck.com/page/deep-learning-framework-character-motion-synthesis-and-editing/}{http://theorangeduck.com/page/deep-learning-framework-character-motion-synthesis-and-editing}.}
We pooled this
%the Edinburgh locomotion
dataset with the locomotion 
%of the database from \cite{holden2015learning,holden2016deep}. We used the locomotion
trials from the CMU \cite{cmu2003mocap}
and HDM05 \cite{muller2007documentation} databases.
We held out a subset of the data with a roughly equal amount of motions in different categories (such as walking, running, and sidestepping) for evaluation, and used the rest for training.
For the animal motion, we used the 30 minutes of dog motion capture from \citet{zhang2018mode}, excluding clips on uneven terrain.
Quadrupedal locomotion allows more gaits
%(e.g., walk, pace, trot, and canter)
than bipedal locomotion (see \citet{zhang2018mode}), but the data also contains motions like sitting, standing, idling, lying, and jumping.
%, again on a flat surface.
We held out two sequences comprising 72 s of data.

Both datasets were downsampled to 20 frames per second and sliced into fixed-length 4-second windows with 50\% overlap for training.
The lowered frame rate both reduces computational demands and decreases over-reliance on autoregressive feedback, as discussed in Sec.\ \ref{ssec:dropout}.
The training data was subsequently augmented by lateral mirroring.
To increase the amount of backwards and side-stepping motion, we further augmented the data by reversing it in time.
This way we obtained 13,710 training sequences from the human data and 3,800 from the animal material.
Preliminary comparisons indicated that the reverse-time augmentation substantially improved the naturalness of synthesised motion.
\begin{table*}[t!]
%\vspace{-\baselineskip}
\centering
\caption{Overview of system configurations considered in this paper.
\new{Numbers with \man{} pertain only to the human model, \dog{} to the dog.} %
%MA was trained on a different machine than other systems.
%``Unk.'' signifies unknown.
%Note that QN trains two separate networks with different loss functions.
}
\label{tab:systems}
\vspace{-1.2ex}
\small{%
\begin{tabular}{@{}lll|ccllll|cc|lllc@{}}
\toprule
 &  &  & Proba- & Task- & Algo. & Context & Hidden & Pose & \multicolumn{2}{l|}{Num.\ params.} & Training\ldots{} &  &  & \tabularnewline
 & Configuration & ID & bilistic? & agnostic? & latency & frames & state & dropout & Man & Dog & Loss func. & Epochs & Time & GPUs\tabularnewline
\midrule 
\multirow{4}{*}{\begin{turn}{90}
Baselines
\end{turn}} & Plain LSTM & RNN & \xmark & \cmark & None & - & LSTM & - & \hphantom{0}1M & \hphantom{0}1M & MSE & 40 & \hphantom{0}0.7\man h & 8\tabularnewline
 & \citet{greenwood2017predicting} & VAE & Partially & \cmark & Full seq. & - & BLSTM & - & \hphantom{0}4M & \hphantom{0}4M & MSE+KLD & 40 & \hphantom{0}6.1\man h & 8\tabularnewline
 & \citet{pavllo2018quaternet} & QN & \xmark & \xmark & 1 sec. & - & GRU & - & 10M & - & Angl./pos.+reg. & 2k/4k & 10\man h & 2\tabularnewline
 & \new{\citet{zhang2018mode}} & \new{MA} & \new{\xmark} & \new{\xmark} & \new{1 sec.} & \new{12}  & \new{-} & \new{-} & \new{-} & \new{\hphantom{0}5M} & \new{MSE} & \new{150} & \new{30\dog{} h} & \new{1}\tabularnewline
\midrule 
\multirow{1}{*}{} & MoGlow & MG & \cmark & \cmark & None & 10 & LSTM & \hphantom{0}95\% & 74M & 80M & Log-likelihood & 291\man & 26\man h & 1\tabularnewline
\midrule 
\multirow{3}{*}{\begin{turn}{90}
Ablats.
\end{turn}} & No pose dropout & MG-D & \tqdb & \tqdb & \tqdb & 10 & \tqdb & \hphantom{00}0\% & 74M & - & \tqdb & \tqdb & 26\man h & \tqdb\tabularnewline
 & No pose context & MG-A & \tqdb & \tqdb & \tqdb & 10 & \tqdb & 100\% & 74M & - & \tqdb & \tqdb & 26\man h & \tqdb\tabularnewline
 & Minimal history & MG-H & \tqdb & \tqdb & \tqdb & \hphantom{0}1 & \tqdb & \hphantom{0}95\% & 54M & - & \tqdb & \tqdb & 23\man h & \tqdb\tabularnewline
\bottomrule
\end{tabular}}%
%\vspace{-1ex}
\end{table*}

We used the same pose representation and control scheme as in \citet{habibie2017recurrent}.
Each pose frame $\x_t$ in the data thus comprised 3D joint positions of a skeleton expressed in a floor-level (root) coordinate system following the character's position and direction.
The root motion was calculated by \newer{Gaussian}-filtering the horizontal, floor-projected hip motion from the original data, which yielded a $(x,\,z)$ trajectory on the ground together with the up ($y$) axis rotation.
The filtering is essential for generalising the synthesis to smooth control signals as provided by an artist or from game-pad input.

The human data had 21 joints ($D=63$ degrees of freedom), while the dog data had 27 joints ($D=81$ degrees of freedom).
This was supplemented \vspace{-0.00001ex}with the frame-wise delta translation and delta rotation (around the up-axis) of the root, which together
% encode the movement of the root node through space.
constitute the control signal $\ctrl_t\in\real^3$ for each frame.
The trajectory of the root over time is computed from the control signal $\ctrl_t$ using integration, and is therefore completely determined by the sequence of control inputs $\ctrl$.
The end result is that the root is constrained to exactly follow a specific path
%through space, like a bead on a stiff wire
on the ground and path-following is essentially perfect; the task of the motion-synthesis model is to generate a sequence of body poses that are consistent with motion along this trajectory.
%We used the 63-dimensional skeleton pose as the output , treating the three displacement variables as the control $\ctrl_t$.
%In other words, the control signal defines a planar track $\ctrl$ through space, along which motion occurs with a given speed and rotation; the models are supposed to complete the motion with an appropriate sequence $\x$ of body poses.
Each dimension in the data and control signal was standardised to zero mean and unit variance over the training data prior to training.%

%\new{Although our main experiments were performed on joint positions in Cartesian coordinates, our method is not specific to any particular pose representation.
%The supplementary video shows that MoGlow also can generate convincing motion also when trained on joint rotations represented using the exponential map, like in \citet{alexanderson2020style}.
%Importantly, this allows applying MoGlow to skinned characters, as seen in the video.
%An alternative route to obtain skinned characters is to train on joint positions in a skeleton with virtual joints like in \citet{smith2019efficient}, and then apply inverse kinematics to recover joint angles, but we would recommend against this approach on the basis that directly outputting joint rotations is lower-dimensional, simpler, and gives competitive results.}

\subsection{Proposed model and ablations}
\label{ssec:models}
We trained the same PyTorch implementation%
\footnote{Please see our project page \href{https://simonalexanderson.github.io/MoGlow/}{https://simonalexanderson.github.io/MoGlow} for links to code, data, and hyperparameters from the evaluation, as well as updated hyperparameter settings that we think further improve output quality.}
%\footnote{Our code was based on \href{https://github.com/chaiyujin/glow-pytorch}{https://github.com/chaiyujin/glow-pytorch}. %
%We will release our MoGlow implementation on GitHub upon paper acceptance.}
of MoGlow on both the human and the animal data.
We used a $\tau=10$-frame time window (0.5 seconds) with $N=16$ steps of flow.
The neural network in each coupling layer comprised two LSTM layers (512 nodes each), followed by a linear (for $\bb{t}_n$) and sigmoid (for $\bb{s}_n$) output layer.
%a fixed-length model as in Sec.\ \ref{ssec:seqcontrol} with $L=4$ blocks of $K=32$ steps of flow each, %splitting $\z_t$-vectors but concatenating $\ctrl_t$-vectors between each block.
%not splitting (but squeezing) $\ctrl_t$-vectors between each block.
%The neural network in the affine coupling layers used two hidden convolutional layers with 512 nodes and ReLU nonlinearities.
%We also trained an autoregressive MoGlow model as in Sec.\ \ref{ssec:conditionalar}, 
%The autoregressive frame dropout rate was set to 0.95 and the time window to $\tau=10$ frames (0.5 seconds).
Model parameters were estimated by maximising the log-likelihood of the training-data sequences using Adam \cite{kingma2015adam} for 160k steps (human) or 80k (quadruped) with batch size 100.
Both models used a learning rate of $10^{-4}$, but for the quadruped we used the Noam learning rate scheduler \cite{vaswani2017attention} with 1k steps of warm-up and peak learning rate $10^{-3}$.
%this was linearly increased to $10^{-3}$ during a warm-up period of 1k steps, and then decayed exponentially so as to fall back towards the minimum in approximately 100k steps.
The autoregressive frame dropout rate was set to 0.95 during training (no dropout was used during synthesis).
%Despite the larger number of steps, the latter model required less wall-clock time due to its smaller size. %We note that training consistently ``just worked'' without the need for tuning optimiser hyperparameters, in contrast to theoretical \cite{mescheder2018training} and practical \cite{lucic2018gans} issues identified with many GAN training paradigms.
We denote this system ``MG'' for MoGlow.
While many GANs and normalising-flow applications heuristically reduce the temperature (standard deviation) of the latent distribution $\Z_t$ at generation time, we found this to be unnecessary, and in fact detrimental to the visual quality of motion sampled from the system.

% TODO: Add figure showing learning curves

To assess the impact of important design decisions, we trained three additional versions of the MoGlow architecture on the human data.
In these, specific components had been disabled from the full MG system:
The first ablated configuration, ``\mbox{MG-D}'' (for ``minus dropout'') turned off data dropout by setting the dropout rate to zero.
As discussed in Sec.\ \ref{ssec:dropout}, we expect this system to exhibit poor adherence to the control signal and establish the utility of introducing data dropout.
The second, ``\mbox{MG-A}'' (for ``minus autoregression''), instead increased the dropout rate to 100\%, thereby completely disabling autoregressive feedback from recent poses $\xrange{t-\tau}{t-1}$.
%(Note that, since the coupling-layer networks $A_n$ take partial information from previous $\z_t$-values as input, and these $\z_t$-values are deterministically related to the output pose $\x_t$, some information about previous poses can still flow through the recurrent connections in the couplings.
%Using 100\% pose dropout therefore does not completely break the sequence model.)
We expect the contrasts between MG and MG-A to show the utility of the autoregressive feedback in the model.
%, and (together with MG-D) demonstrate that the optimal pose dropout rate is greater than 0 but less than 100\%.
The final ablation, ``\mbox{MG-H}'' (for ``minus history'') changed $\tau$ from ten frames (0.5 s of history information) down to a single frame.
This is the minimum history length at which the model remains autoregressive; any pose or control information older than $t-1$ must now be propagated by the LSTMs in $A_n$ instead.
(Unlike MG-D and MG-A, MG-H also affects the control information, in addition to the autoregressive feedback.)
We expect this ablation to demonstrate the utility of providing the flows with an explicit memory buffer of the most recent pose and control inputs, in addition to the long-range information about past inputs propagated through the recurrent hidden state.
Table \ref{tab:systems} summarises the properties and training of the proposed system and its ablations.

\subsection{Baseline systems}
\label{ssec:baselines}
To put the performance of MoGlow in perspective, we compared against a number of other motion generation approaches.
The first of these is held-out motion capture recordings, which we label ``NAT'' for natural.
(We prefer not to use the term ``ground truth'', since there is no one true way to perform a given motion.)
These motion examples function as a top line.

We also compared against two task-agnostic motion-synthesis approaches, labelled ``RNN'' and ``VAE''.
The first of these, RNN, is a deterministic system that maps control signals $\ctrl_t$ to poses $\x_t$ using a standard unidirectional LSTM network (one hidden layer of 512 nodes followed by a linear output layer) and was trained to minimise the mean squared error (MSE).
Because our path-based control signal does not suffice to disambiguate the motion, we expect this generic method to exhibit considerable regression to the mean, for instance visible through foot-sliding.
This is emblematic of task-agnostic deterministic methods.
The other task-agnostic baseline, VAE,
%For purposes of comparison we also considered three baseline motion-generation methods in our experiments: ``RNN'', ``VAE'', and ``QN''. 
%The RNN system was a standard unidirectional LSTM network, one layer (512 nodes), followed by a linear output layer, to map control signals $\ctrl_t$ to poses $\x_t$.
%VAE
is a reimplementation of the conditional variational autoencoder architecture used for speech-driven head-motion generation in \citet{greenwood2017predicting,greenwood2017predictingb}, but in our case predicting motion $\x$ from $\ctrl$.
%Some architectural choices are not easily gleaned from the papers
We used encoders and decoders with two bidirectional LSTM layers (256 nodes each way) and a linear output layer.
The encoder used mean-pooling to map to a latent space with two dimensions per sequence.
% a 2D per-sequence latent space.
%We used an encoder with two bidirectional LSTM (BLSTM) layers (256 nodes each way) mean-pooled into a 2D latent space, followed by a decoder with two BLSTM layers (256 nodes).
Due to the bidirectional LSTMs in the conditional decoder, interactive control is not possible with this approach.
%The and trained to minimise the (expected) MSE of decoder predictions together with the standard KL-divergence prior of variational autoencoders.
Unlike the RNN baseline, VAE represents a partially probabilistic model, which should enable it to cope with motion that is random and ambiguous also when conditioned on the control signal.
The model does not incorporate any assumptions specific to head-motion data, and can be considered representative of the state of the art in probabilistic, task-agnostic motion generation.
We say that this system is ``partially probabilistic'' since the decoder is trained to minimise the MSE and treated as deterministic rather than stochastic at synthesis time.
%In particular, the deterministic MSE loss function for the decoder is equivalent to assuming an isotropic Gaussian output distribution from the decoder.
%However, as is common in generative VAE applications, output was produced deterministically from the mean of the (implied Gaussian) decoder output distribution, $\widehat{\x}=\mean\left\[p\given{\x}{\z}\right\]$, instead of actually sampling from $p\given{\x}{\z}$ by adding Gaussian noise to $\widehat{\x}$.
%However, the isotropic Gaussian noise that should be added to the decoder output when generating from this probabilistic interpretation is ignored, with the only randomness in output data coming from the two-dimensional per-sequence latent variable $\z$.
As a consequence, output samples from the system have artificially reduced randomness compared to sampling from the full probabilistic model described by the fitted VAE, whose decoder is a Gaussian distribution.
Such reduced-entropy generation procedures are common in practice since they tend to improve subjective output quality (see Sec.\ \ref{ssec:probmotion}), but also indicate that the underlying model has failed to convincingly model the natural variation in the data.
%While this reduced-entropy generation procedure might be subjectively more appealing (see Sec.\ \ref{ssec:probmotion}) and is standard in most VAE applications, it does not attempt to address the central challenge of building an accurate probabilistic model of motion whose samples are both inherently realistic and diverse.
\begin{figure}
\centering
%\hfill
\includegraphics[width=0.48\columnwidth]{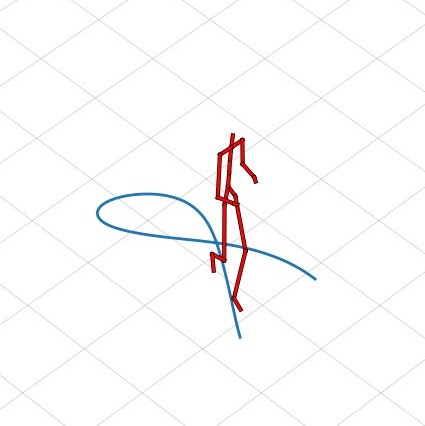}
\hfill
\hfill
\includegraphics[width=0.48\columnwidth]{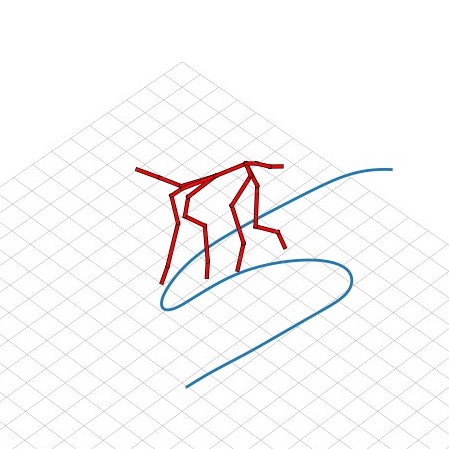}
%\hfill
\vspace{-0.5em}
\caption{Still images cropped from videos of MG output. %
The path followed by the root node, which is completely determined by $\ctrl$, is visualised as a blue curve projected onto the ground plane.}
\label{fig:skeleton}
\vspace{-1em}
\Description{Two images of red stick figures, one human and one dog, walking along a blue path on an isometric grey grid visualising the ground plane.}
\end{figure}

Finally, we also compared our proposed method with a leading task-specific system \new{in each of the two domains}.
%for the task of human locomotion generation.
Human locomotion generation, to begin with, is a mature field where many approaches may be considered state-of-the-art.
One example is the recently proposed QuaterNet \cite{pavllo2018quaternet}, which we included in our evaluation as system ``QN''.
%QN refers to QuaterNet \cite{pavllo2018quaternet} 
In order not to compromise QN motion quality, we used the code, hyperparameters, and control scheme made available by the original QuaterNet authors.%
\footnote{Please see \href{https://github.com/facebookresearch/QuaterNet/}{https://github.com/facebookresearch/QuaterNet}.}
This introduced a number of minor differences compared to other systems.
Specifically, the QuaterNet reference implementation contains a number of preprocessing steps that change the motion:
First the input path is approximated by a spline, and facing information and local motion speed are replaced.
This control scheme causes the character to always face the direction of motion, preventing sidestepping or walking backwards.
Short spline segments are then lengthened, preventing the model from standing still.
%We hypothesise that this avoids making tight corners, which might look unrealistic when the character is constrained to always face the instantaneous direction of travel.
One goal with MoGlow is to deliver high-quality motion without such custom, task-specific processing steps.
Finally, we resampled the output from the trained QN system to 20 fps to match the other systems in the evaluation.

\new{For the quadruped locomotion task, we compared with the mode-adaptive neural networks from \citet{zhang2018mode}.
%As they used the same data as us,
Since they trained on the same dataset as us, we used their pretrained model%
\footnote{\new{Available at \href{https://github.com/sebastianstarke/AI4Animation/}{https://github.com/sebastianstarke/AI4Animation}.}}
as our system ``MA'' for best results.
%which we included as system ``MA''.
%Provide a pre-trained model at, which we made use of for best results.
To our knowledge, no data was held out from their training.
In the absence of held-out control signals, MA was therefore only evaluated on synthetic control input.
For the experiments we set the MA style input to ``move'' and the correction parameter $\tau$ to 1, to make the model follow the input patch exactly, like the other systems in the evaluation.
%\newer{The style-input of the net was set to ``move'' since we evaluated locomotion.}
MA output was also resampled to 20 fps.}

%QN differs from other methods considered in that the figure always faces the direction of motion; side-stepping etc.\ is not possible, and it takes 1.0 seconds of future control information into account (at the expense of introducing substantial latency into the control).
%Also, while the two first baselines (RNN and VAE) are general methods that do not assume that the motion is periodic, the QN implementation contains assumptions specific to bipedal locomotion.
%We therefore did not train a QN system on the quadruped data.

In summary, RNN and VAE are task-agnostic systems -- one deterministic, one probabilistic -- while QN \new{and MA instead represent the task-specific state of the art in their respective task.}
%exemplifies the task-specific state of the art for human locomotion generation (and was not applied to the quadruped data).
We note that, unlike RNN and the MG systems, VAE\new{, QN, and MA are} noncausal, in the sense that \new{their output} depends on future control-input information.
We expect this ability to ``see the future'' to benefit the quality of the motion generated for these systems, but it comes at the cost of introducing algorithmic latency, preventing the type of responsive control that MG allows.
All our models were trained on a system with 8 Nvidia 2080Ti GPUs.
An overview of the different systems, including information such as training time, model size, and the number of GPUs used, is provided in Table \ref{tab:systems}.
%Increased training time or model size did not noticeably improve output quality for the baselines.

\section{Results and discussion}
\label{sec:results}
This section details our subjective (Secs.\ \ref{ssec:subjective} through \ref{ssec:analysis}) and objective (Sec.\ \ref{ssec:objective}) evaluations of the different motion-generation methods \new{from Sec.\ \ref{sec:experiment}}, and how we interpret the results.
\new{We then describe (Sec.\ \ref{ssec:exploration}) experiments that explore the probabilistic aspects of the model, and consider its use beyond locomotion.
We then conclude with a discussion of drawbacks and limitations (Sec.\ \ref{ssec:limitations})}.

\subsection{Subjective evaluation setup}
\label{ssec:subjective}
Since our goal is to create lifelike synthetic motion that appears convincing to human observers, subjective evaluation is the gold standard.
To this end we conducted \new{several user studies} to measure motion quality on the \new{two tasks}.
The stimuli used in both studies were short
%(4-second)
animation clips where motion was visualised using a stick figure
%(a so-called \emph{skeleton})
seen from a fixed camera angle; see Fig.\ \ref{fig:skeleton}.
A curve on the ground marked the path taken by the figure in the clip.
Clips were generated for all systems in Table \ref{tab:systems} and from held-out motion-capture recordings (``NAT'').
%(We prefer not to use the term ``ground truth'', since there is no one true way to perform a given motion.)
For MG, one second of preceding motion was pre-generated before the four seconds that were displayed and scored, to remove the effects of motion initialisation.
Since the QuaterNet preprocessing changes the motion duration, the segmentation points for the evaluation clips (and also the camera azimuth) differ between QN and the other systems.
%, e.g., video \texttt{QN\_13} in the supplement approximately matches \texttt{MG\_12}.

In addition to motion generated from held-out natural control signals \new{(20 human, 8 dog)}, the evaluation also included synthetic control signals \new{(7 human, 10 dog)} \new{with a range of motion speeds and directions}, for which no natural counterpart was available.
%These were created by keeping $\ctrl_t$ constant within each clip, resulting in motions like standing still and walking and running forwards, backwards, sideways, and in a circle.
Generalising well to synthetic control is important for computer animation, video games, and similar applications.
%However, methods reliant on receiving footfall cues from the control input (e.g., residual periodicity in the root-node motion) can be expected to degrade on such data.

Evaluation participants were recruited using the \emph{Figure Eight} crowdworker platform at the highest-quality contributor setting (allowing only the most experienced, highest-accuracy contributors).
For each clip, participants were asked to grade the perceived naturalness of the animation on a scale of integers from 1 to 5, with 1 being \emph{completely unnatural} (motion could not possibly be produced by a real person\new{/dog}) and 5 being \emph{completely natural} (looks like the motion of a real person\new{/dog}).
Every system in Table \ref{tab:systems} had one stimulus generated for every control signal considered, \new{with a few exceptions:} QN was not applied to synthetic control signals, since these contained a large fraction of control inputs involving walking sideways, backwards, and standing still, motion that the QN reference implementation from \citet{pavllo2018quaternet} cannot perform (instead rendering these as forwards motion).
\new{MA was not applied to our natural test inputs, since these were not held out from MA training.
The ablated systems were only evaluated on the human locomotion task.}
%a relatively large amount of backwards and sideways motion which the QN implementation from \citet{pavllo2018quaternet} was unable to faithfully recreate (instead rendering these as forwards motion).
This yielded a total of 202 \new{human} animations being evaluated (160 with held-out control and 42 with synthetic control) \new{and 72 dog animations (32 held-out, 40 synthetic control)}.
The order of the animation clips was randomised, and no information was given to the raters about which system had generated a given video, nor about the number of systems being evaluated in the test.

Interspersed among the regular stimuli were \new{a handful of}
%5 instances each of 5
clips with deliberately \emph{bad} animation taken from early iterations in the training process (labelled ``BAD'').
These were added as ``attention checks'' to be able to filter out unreliable raters: Any rater that had given any one of the BAD animations a rating of 4 or above, \emph{or} had given any of the NAT clips a rating below 2, was removed from the analysis.
Ratings that were too fast (the rater replied before the video had finished playing) were also discarded.
Prior to the start of the rating phase, participants were trained by viewing example motion videos from the different conditions evaluated, as well as some of the bad examples mentioned above.
Motion examples can be seen in \href{https://youtu.be/pe-YTvavbtA}{our presentation video}
%at \href{https://youtu.be/ozVldUcFjZg}{https://youtu.be/ozVldUcFjZg}.
and in the supplementary material, which contains all video clips from the subjective evaluation.%
%\begin{figure}
%\centering
%\includegraphics[width=1\columnwidth]{fig_bar_ratings}
%\caption{Mean ratings of the clips in the subjective evaluation, with 95\% confidence intervals.}
%\label{fig:ratings}
%\Description{A bar chart with confidence intervals of different systems with held-out and synthetic control. MG and QN are almost indistinguishable from NAT in the held-out control case. MG does not degrade much on the synthetic control, although the confidence interval is wider there since fewer such videos were evaluated.}
%\end{figure}%
\begin{table}[t!]
%\vspace{-\baselineskip}
\centering
\caption{\new{Mean subjective ratings with confidence intervals.
Significant differences from MG are indicated by ** ($p<0.01$) and * ($p<0.05)$.}}
\label{tab:subjective}
\vspace{-1.2ex}
\small{%
\begin{tabular}{@{}l|cc|cc@{}}
\toprule
 & \multicolumn{2}{c|}{\new{Human}} & \multicolumn{2}{c}{\new{Quadruped}}\tabularnewline
ID & Held-out $\ctrl$ & Synthetic $\ctrl$ & \new{Held-out $\ctrl$} & \new{Synthetic $\ctrl$}\tabularnewline
\midrule 
NAT & 4.27$\pm$0.11\hphantom{**} & - & \new{4.25 $\pm$ 0.06**} & \new{-}\tabularnewline
\midrule 
RNN & 3.10$\pm$0.15** & 1.9$\pm$0.2** & \new{2.81 $\pm$ 0.10**} & \new{1.14 $\pm$ 0.04**}\tabularnewline
VAE & 3.95$\pm$0.13\hphantom{**} & 3.1$\pm$0.3** & \new{3.55 $\pm$ 0.08\hphantom{**}} & \new{2.14 $\pm$ 0.20**}\tabularnewline
QN & 4.21$\pm$0.10\hphantom{**} & - & \new{-} & \new{-}\tabularnewline
\new{MA} & \new{-} & \new{-} & \new{-} & \new{3.78 $\pm$ 0.10\hphantom{**}}\tabularnewline
\midrule 
MG & 4.17$\pm$0.11\hphantom{**} & 4.0$\pm$0.2\hphantom{**} & \new{3.71 $\pm$ 0.18\hphantom{**}} & \new{3.57 $\pm$ 0.20\hphantom{**}}\tabularnewline
\midrule 
MG-D & 3.66$\pm$0.16** & 2.1$\pm$0.2** & \new{-} & \new{-}\tabularnewline
MG-A & 2.86$\pm$0.16** & 3.2$\pm$0.3** & \new{-} & \new{-}\tabularnewline
MG-H & 3.87$\pm$0.13*\hphantom{*} & 3.9$\pm$0.3\hphantom{**} & \new{-} & \new{-}\tabularnewline
\bottomrule
\end{tabular}}%
\vspace{-1em}
\end{table}

\subsection{Analysis and discussion of subjective evaluation}
\label{ssec:analysis}
A total of \new{645} raters \new{(296 human data/349 dog data)} participated in the evaluation, of which \new{89 (49/40)} were removed as unreliable (see above). In total, \new{10,355} ratings were collected \new{(5,083/5,272)}.
1,533\new{/983} of these were discarded due to unreliable rater (1,344\new{/813}) or too fast response time (189\new{/170}), resulting in a total of 3,550\new{/4,289} ratings across 227\new{/80} clips being evaluated (\new{both regular and BAD}), amounting to between \new{8 and 60}
%8 and 19
ratings per stimulus.
The mean scores for each system configuration and control-signal class are
%graphed in Fig.\ \ref{fig:ratings} and
tabulated in Table \ref{tab:subjective}.

\new{For the human motion, a} one-way ANOVA revealed a main effect of the naturalness rating ($F=223$, $p < 10^{-288}$).
A post-hoc Tukey multiple-comparisons test was applied in order to identify significant differences between conditions ($\text{FWER}=0.05$).
For the held-out control conditions, MG was rated significantly higher than RNN and all ablations.
For the synthetic control conditions, MG was rated significantly higher than all other systems except the ablation system MG-H.  
\new{The same analysis for the quadruped motion again revealed a main effect of the naturalness rating ($F=172$, $p < 10^{-100}$ for held-out $\ctrl$, $F=803$, $p < 10^{-296}$ for synthetic).
The post-hoc Tukey multiple-comparisons test revealed significant differences between MG and all other systems, except between MG and VAE on the held-out control and between MG and MA on the synthetic control.
%synthetic control conditions except between MG and MA.
%For the held-out control conditions there were significant differences between all but MG and VAE
}
95\%-confidence intervals for the mean scores based on \new{these analyses} are included in
%Fig.\ \ref{fig:ratings} and 
Table \ref{tab:subjective}, which also indicates significant differences between MG and other systems.

Among the task-agnostic methods in the experiment, MG substantially outperforms both RNN and VAE.
Despite \new{these MG systems} being trained to predict joint positions rather than joint rotations, \new{they are} seen to respect constraints due to bone lengths, ground contacts, etc.
Furthermore, the \new{rated motion quality} of MG \new{on each task} is comparable to the \new{respective} task-specific state of the art \new{(the difference between MG and either QN or MA is not statistically significant), and comes within 0.1 points of natural motion for the biped.}
%Clearly, the two best performing models in the experiment are QN and MG, which
%\new{it} consistently generate\new{s} visually-convincing motion and come\new{s} within \new{0.1 or 0.2 points of the task-specific state of the art}.
This is despite \new{the task-specific systems} having a full second of algorithmic latency,
%(providing additional information about upcoming motion)
while MG \new{is task-agnostic and} has none.
\newer{We note that stimuli where the root is completely still are generally rated lowest for MG and MA, and not possible to generate with QN.}
%\newer{We note that for both tasks, synthetic MG motion where the root is completely still is ranked the lowest.
%This is also true for MA on the quadruped task.}%
%\newer{For both tasks, the lowest-rated MG stimuli are those where the root is completely still, where model output is seen to nonetheless exhibit a small amount of limb motion.
%The same observations hold for MA on the dog data.%
%In these cases, model output is seen to exhibit a small amount of limb motion.%
%}%
%(Interestingly, NAT only reached 4.27 on the five-point rating scale, which might reflect the inherent unnaturalness of the stick-figure representation.)
%QN might have a slight edge on MG in terms of quality because it uses a full second of future control information during synthesis -- unlike MG, which has no algorithmic latency in the control -- and because it is unable to sidestep or walk backwards, instead rendering these as forwards motion, which may be perceived by the raters as an a-priori more natural direction of motion.

Among other results, the performance of the ablations MG-D and MG-A versus the full MG system indicate that both autoregression and data dropout are of great importance for synthesising natural motion.
A longer memory length
%for $\xrange{t-\tau}{t-1}$ and $\crange{t-\tau}{t-1}$
of $\tau=10$ frames for MG, compared to $\tau=1$ for MG-H, also benefited the model.
It can be observed that RNN, VAE, and MG-D quality degrades substantially on synthetic control signals, creating a highly significant difference with respect to MG.
We hypothesise that this, for MG-D, is due to artefacts of poor control without data dropout (such as running in place; see Sec.\ \ref{ssec:dropout}), and, for RNN and VAE, due to the systems being dependent on footfall cues (e.g., residual periodicity in the root-node motion) not present in the synthetic motion control.
The full MoGlow model, in contrast, generalises robustly to synthetic control signals.

%\new{We note that many methods in the literature primarily demonstrate results on rapid motion such as sprinting (cf.\ \citet{ling2020character}).
%MG achieves high scores in our evaluation because it is able to do well on a range of different motion speeds, including slow movement where foot sliding is easier to perceive.}
%\begin{figure}
%\centering
%\includegraphics[width=0.8\columnwidth]{fig_footsteps}
%\caption{Footstep estimation plotted against velocity tolerance for all held-out sequences in the human data.}
%\label{fig:footsteps}
%\end{figure}

%condition were 4.05 for \textbf{NAT}, 3.68 for \textbf{MG}, and 3.66 for \textbf{FL}.
%(For comparison, the \emph{bad example} control condition received a mean score of 1.34, however these data points were not used in the rest of the analysis.)
%Fig.\ \ref{fig:scoreshist} reveals the distribution of responses for the different conditions. It can be noted that the most common rating for all three conditions was 5.

%While the animations produced by the two systems in the experiment were perceived somewhat less natural than the ground-truth motion, it is interesting to note that the quality of the output from the proposed autoregressive system is rated on par with that from the noncausal, fixed-length implementation. The main implication of this is that the advantages of the autoregressive method, i.e., the ability to generate animation continuously on the fly -- enabling applications such as real-time control of computer game-characters, robots, or virtual actors -- can be gained without sacrificing motion quality. 

\subsection{Objective evaluation}
\label{ssec:objective}
\begin{figure}[!t]
\centering
\includegraphics[width=0.4985\columnwidth]{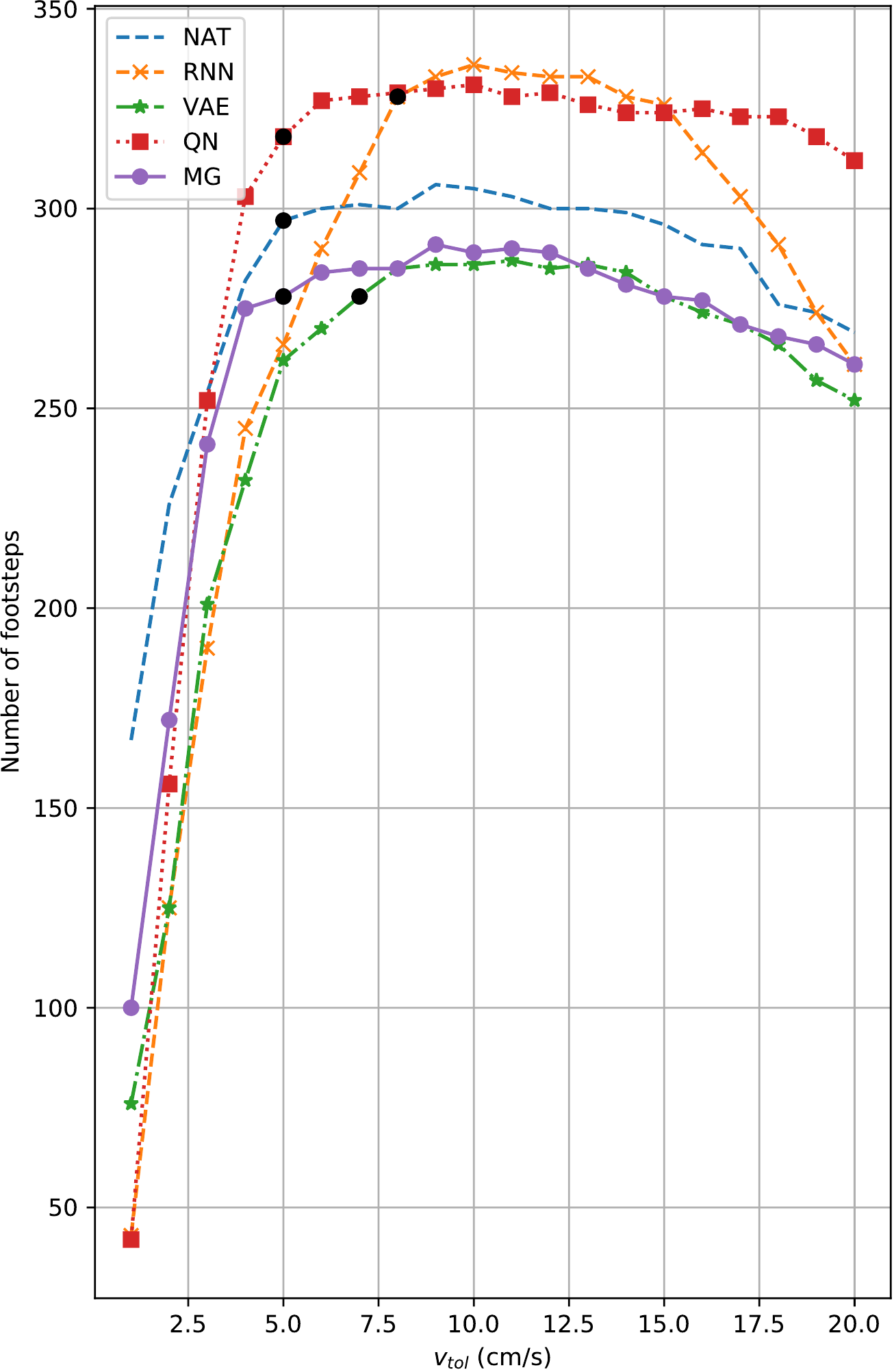}
\hfill
\includegraphics[width=0.48\columnwidth]{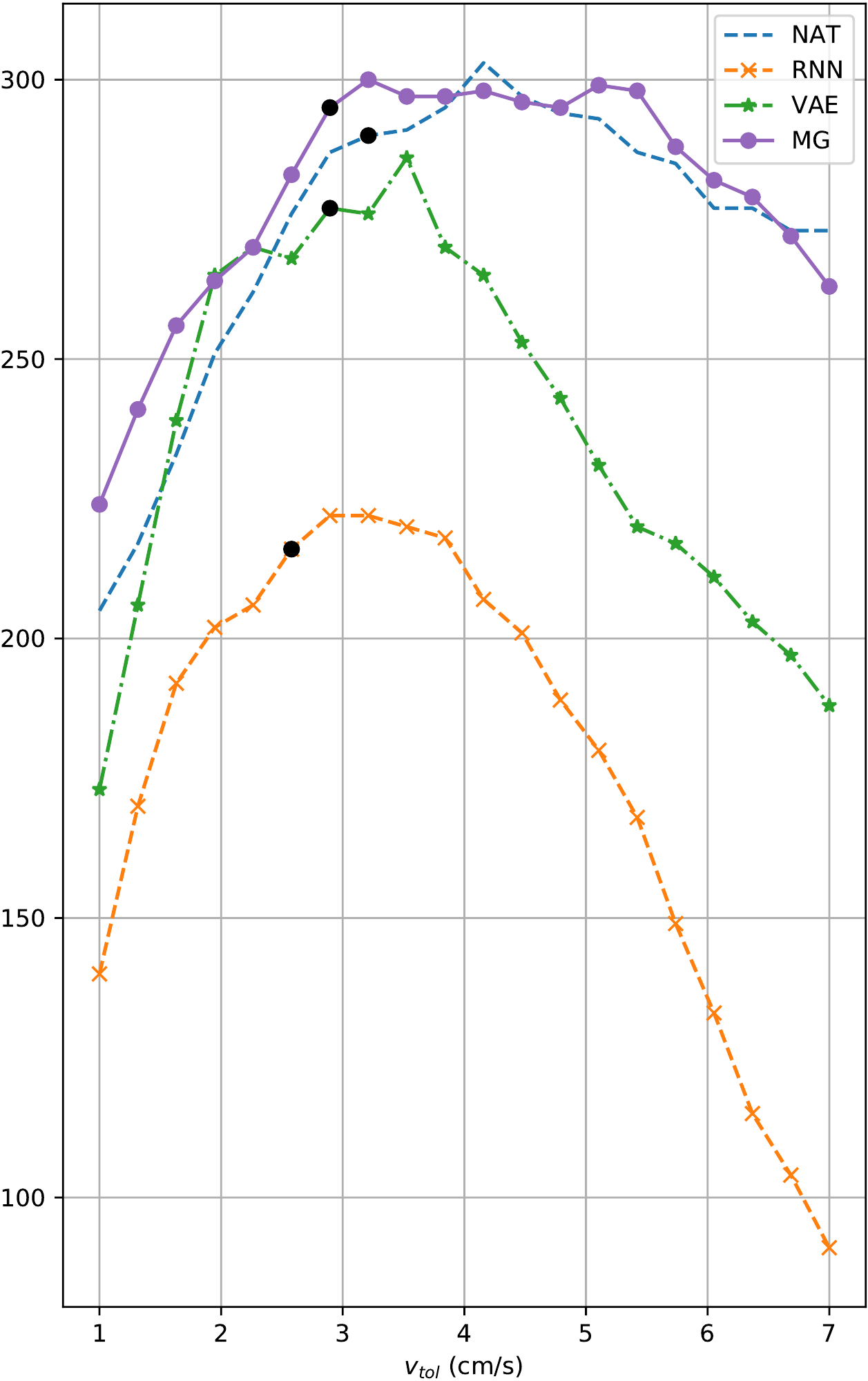}
%\begin{center}
%\includegraphics[width=0.9\columnwidth]{foot_steps_man-crop}\\
%\vspace{1.5ex}
%\includegraphics[width=0.9\columnwidth]{foot_steps_dog-crop}
%\end{center}
\vspace{-1.2ex}
\caption{Footstep count $f_{\mathrm{est}}$ as a function of speed tolerance $v_{\mathrm{tol}}$ (cm/s) for the human (left) and quadruped (right) datasets.
Black dots identify locations used to determine $v^{(95)}_{\mathrm{tol}}$ for each curve.}
\label{fig:footsteps}
\vspace{-3ex}
\Description{Two plots, one for each data source, of per-system curves that rise, then plateau, and then eventually decrease again. The characteristics of the individual system curves are described in more detail in the article text and in Table 3.}
\end{figure}
%that possibly of the pervasive and noticeable artefacts, namely foot sliding, can be quantified objectively with relative ease.
%In general, it is difficult to perform objective evaluation of plausible motion synthesis as the aim is not to directly reproduce the ground truth but rather to generate equally probable motion. Thus, direct comparison with low-level properties such as frame-wise joint positions are not informative. Instead, we base our objective evaluation on the analysis of feet contacts. Feet contacts have physical constraints that can be objectively measured, and foot-sliding artefacts are particularly known to affect the perceived naturalness of animated locomotion.

%Similar to \citet{holden2016deep,holden2017phase,pavllo2018quaternet}, we base our footstep detection on a heuristic approach. Specifically, we
Given the salience and importance of foot-sliding artefacts in locomotion synthesis, we base our objective evaluation on footstep analysis, with footsteps estimated as time intervals where the horizontal speed of the heel joints (bipeds) or toe joints (quadrupeds) are below a specified tolerance value $v_{\mathrm{tol}}$.
At low values of $v_{\mathrm{tol}}$, many ground contacts exhibit too much motion (due to foot sliding or motion-capture uncertainty), and are not classified as steps.
%Plots of $f_{\mathrm{est}}$, the estimated number of footsteps, as a function of $v_{\mathrm{tol}}$ therefore typically show a
As the tolerance is increased, the number of footsteps identified, $f_{\mathrm{est}}$, first rises but then quickly plateaus at a static maximum value representing the total number of footsteps in the sequence.
A model that produces foot-sliding artefacts will require higher tolerance before reaching its maximum.
If the tolerance is increased further, the estimated number of footsteps eventually begins to decrease as separate footsteps start to be merged.

Plots of $f_{\mathrm{est}}$ as a function of $v_{\mathrm{tol}}$ \new{on held-out data} are provided in Fig.\ \ref{fig:footsteps}; the human and dog motion clips used as the basis for these plots and for the associated analysis are available in the supplement.
\new{(MA is not included since no data was held out from its training.)}
The plots show that MG is able to stay close to NAT in both scenarios.
QN, which only is available for the human data, generates slightly too many steps, but is otherwise close to the natural footstep profile.
The quadruped data appears to be more challenging than the human data, with the peaked behaviour of the estimated number of footsteps $f_{\mathrm{est}}$ for RNN and VAE indicating less distinctive synthetic locomotion that is likely to exhibit substantial foot sliding.
MG, in contrast, again shows an $f_{\mathrm{est}}$-profile very similar to that of natural motion.

For each model, we incremented $v_{\mathrm{tol}}$ in small steps (1.0 cm/s for human, 0.3 cm/s for quadruped) and extracted the first tolerance value $v^{(95)}_{\mathrm{tol}}$ that reached 95\% of the maximum number of footsteps identified for that model in our evaluation.
These points are shown as black dots on the curves in Fig.\ \ref{fig:footsteps}.
%The lowest speed tolerance at which 95\% of the total steps are captured, here dubbed $v^{(95)}_{\mathrm{tol}}$
%This number gives an indication of the degree of foot sliding in the motion; the lower the better.
The tolerance threshold $v^{(95)}_{\mathrm{tol}}$ essentially measures the 95th percentile of foot sliding in the motion.
The lower this is, the crisper the motion is likely to be.

Table \ref{tab:objective} shows the total estimated number of footsteps, the speed threshold, and the mean and standard deviation of the duration of the steps for different systems when resynthesising the held-out data from the two datasets.
We note that MG almost always is the model that most closely adheres to the ground truth behaviour.
Especially interesting is that MoGlow matches not only the mean but also the standard deviation of the natural step durations.
Such behaviour might be expected from an accurate probabilistic model, whereas deterministic models, not having any randomness and thus no entropy, are fundamentally limited not to match the statistics of the natural distribution in all respects.
%We also note that sorting the results for human locomotion on foot-slide yields a model ranking closely resembling the ranking from the subjective evaluation.
%This supports the common notion that foot sliding is the most salient feature for perceived naturalness, meaning that the corresponding objective statistics for the dog data are likely to provide a good indicator of the perceptual motion quality also by the quadrupedal models.
%Plots of $f_{\mathrm{est}}$ across $v_{\mathrm{tol}}$ are provided in 
%Appendix \ref{sec:extendedsteps}.
%the supplementary material, along with videos from all models and datasets.
%the quadrupedal motion sequences upon which the numbers in the second half of Table \ref{tab:objective} are based.
\begin{table}[t!]
%\vspace{-\baselineskip}
\centering
\caption{Results from the objective evaluations: total number of footsteps $f_\mathrm{est}$, speed tolerance $v^{(95)}_\mathrm{tol}$ (cm/s) for capturing 95\% of steps, mean and standard deviation of step durations (s)\new{, and bone-length RMSE (cm)}. %
The number closest to its natural counterpart in each column is shown in bold.}
\label{tab:objective}
\vspace{-1.2ex}
\small{%
\begin{tabular}{@{}l|ccccc|ccccc@{}}
\toprule
 & \multicolumn{5}{c|}{Human} & \multicolumn{5}{c}{Quadruped}\tabularnewline
ID & $f_\mathrm{est}$ & $v^{(95)}_\mathrm{tol}$ & $\mu$ & $\sigma$ & RMSE & $f_\mathrm{est}$ & $v^{(95)}_\mathrm{tol}$ & $\mu$ & $\sigma$ & RMSE\tabularnewline
\midrule 
NAT & 297 & 5.0 & 0.31 & 0.26 & \new{-} & 290 & 3.2 & 0.61 & 0.71 & \new{-}\tabularnewline
\midrule 
RNN & 328 & 8.0 & 0.39 & 0.39 & \new{1.7\hphantom{0}} & 216 & 2.6 & 0.72& 1.05 & \new{2.3\hphantom{0}}\tabularnewline
VAE & \textbf{278} & 7.0  & 0.35 & 0.30 & \new{1.7\hphantom{0}} & 277 & \textbf{2.9} & \textbf{0.61} & 0.90 & \new{2.0\hphantom{0}}\tabularnewline
QN & 318 & \textbf{5.0} & 0.23 & 0.19 & \new{\textbf{0.07}} & - & - & - & - & \new{-}\tabularnewline
\midrule 
MG & \textbf{278} & \textbf{5.0} & \textbf{0.32} & \textbf{0.23} & \new{0.50} & \textbf{295} & \textbf{2.9} & 0.57 & \textbf{0.75} & \new{\textbf{0.51}}\tabularnewline
\bottomrule
\end{tabular}}%
\vspace{-1ex}
\end{table}
%
%5.0
%8.0
%7.0
%5.0
%5.0
%[297        0.31498316  0.26199073]
%[328         0.39771341  0.3975565 ]
%[278         0.35485612  0.29916517]
%[318         0.23490566  0.18528872]
%[278         0.3205036   0.23732721]
%
%3.2105263157894735
%2.5789473684210527
%2.894736842105263
%2.894736842105263
%[290         0.6087931   0.70984476]
%[216         0.72430556  1.05137673]
%[277        0.60722022  0.90653575]
%[295        0.57491525  0.75451044]
%
%
%
%0.6
%0.9999999999999999
%0.8999999999999999
%0.6
%0.6
%[295.           6.17966102   5.19696064]
%[329.           7.76291793   7.75489013]
%[278.           7.10071942   5.98053794]
%[315.           4.5015873    3.65788958]
%[278.           6.21942446   4.69869686]

\new{Since the task-agnostic models in the objective evaluation were trained on joint positions, bone lengths need not be conserved in model output.
This can lead to bone-stretching artefacts, and joints may even fly apart; cf.\ \citet{ling2020character}.
Fortunately, bone-length deviation is easy to quantify objectively.
Table \ref{tab:objective} reports the RMSE of bone length in cm, simultaneously averaged across all joints and time-frames in the test data.
We see that the error is small, meaning that bone lengths in MG output are stable and consistent.}

\subsection{\new{Probabilistic aspects and further experiments}}
\label{ssec:exploration}
\new{Having evaluated motion quality in-depth across tasks, we now present evidence to validate the wide applicability and the probabilistic aspects of the model.
To increase the relevance for computer-graphics applications, we here change the pose representation to joint angles and apply the synthesised motion to a skinned character.
We note that another option for obtaining skinned characters would be to
%apply inverse kinematics to the joint positions
train on joint positions in a skeleton with virtual joints like in \citet{smith2019efficient}, and then apply inverse kinematics to recover joint angles, although this would add another computational step.}

%As MoGlow is not specific to any particular pose representation, these models were trained on joint rotations represented using the exponential map, like in \citet{alexanderson2020style}, and applied to skinned characters.

%Another option for obtaining skinned characters would be to train on joint positions in a skeleton with virtual joints like in \citet{smith2019efficient}, and then apply inverse kinematics to recover joint angles, but we would recommend against this since directly outputting joint rotations is lower-dimensional, simpler, and gives competitive results.}

\new{We created a new MoGlow model designed to investigate the ability of the method to learn from diverse motion data and reproduce its distribution.
\newer{For this model, we constructed a new dataset by pooling}
%the probabilistic aspects.
%To do this we trained a MoGlow model with path control on a dataset created by pooling
%the human locomotion datasets from Sec.\ \ref{ssec:data} with material from
the LaFAN1 dataset from \citet{harvey2020robust}, along with the Kinematica dataset.%
\footnote{\new{The data is available at \href{https://github.com/ubisoft/Ubisoft-LaForge-Animation-Dataset/}{https://github.com/ubisoft/Ubisoft-LaForge-Animation-Dataset} and at \href{https://github.com/Unity-Technologies/Kinematica_Demo/}{https://github.com/Unity-Technologies/Kinematica\_Demo}, respectively.}}
%Our fist pooled dataset, labelled ``Pool1'', included only the locomotion trials from the two new databases, while the second (``Pool2'') included everything from LaFAN1 and Kinematica, save for the climbing and running on walls in the latter.
We excluded trials involving wall and obstacle interaction as well as dancing, falling, stumbling, fighting, and sitting or lying on the ground.
Nonetheless, this new data contains more varied motion than the data from Sec.\ \ref{ssec:data}, including crouching, hopping, walking while aiming, etc.
This yielded a total of \newer{1 h} 
%72k 
of data at 20 Hz (%
%1 h
augmented to 4 h as before).}
\newer{All motion was retargeted to a uniform skeleton and the joint angles were converted to exponential maps \cite{grassia1998practical}.
The hips were expressed local to the floor-projected root, similar to before.
For the new model, data dropout was reduced to 60\%, which proved to generate smooth motion without losing adherence to the control.
During synthesis, the raw model output was applied directly to the character, without any post-processing such as foot stabilisation}.
%We included all material from the trials with climbing includes motion such as crouching, jumping, running while aiming, dancing, stumbling and falling, etc.
%on the LaFAN1 database\footnote{.} from , which contains 497k 30 Hz mocap frames from diverse motion, not only simple locomotion but also jumping, falling, etc.

\new{As shown in our presentation video and in Fig.\ \ref{fig:randomness}, we find that MoGlow not only is able to learn to produce high-quality motion from the new data, but that model output also successfully reflects the diversity of the material, and random samples of motion along the same path may take very different forms.
MoGlow can thus produce a wide gamut of different motions for fixed control input, as expected for a strong probabilistic model under weak control signals.
This is beneficial for increasing variation and naturalness, for example automatically generating sniffing behaviour when the dog is moving slowly.
%, we see the MG dog sniff around the ground in some sequences.
%This can, for example, be beneficial for generating idle animations, in the same way the MG dog tends to sniff around the ground when standing still.
By training a similar model on all the human motion capture material, with no trials except climbing and running on walls excluded, even more varied output was produced, as shown at the very end of our presentation video.
%(We also trained one final model on a very large dataset of human motion where no data except climbing and running on walls was excluded; some samples from this model are included in the presentation video.)
%Control over this diversity may be obtained through style-control techniques like those described in \citet{alexanderson2020style}, or by reducing sampling temperature.

In situations where greater control over motion diversity is desired, this may be obtained by reducing the sampling temperature or by using other, stronger control signals.
For example, crouching or crawling motion might be consistently recovered without manual annotation of training data by training models where pelvic distance above ground is a control input instead of a model output.}

\new{Nothing about MoGlow is specific to locomotion.
%To further show that MoGlow is not specific to locomotion, our supplementary material provides video examples of a MoGlow model trained to generate orchestra conductor motion from motion capture and acoustic features from a few performances of classical music.
The generality of the approach is demonstrated by follow-up work \cite{alexanderson2020style}, performed after the locomotion studies described in this article but published before this article appeared, that shows that MoGlow successfully generalises to synthesising speech-driven gesture motion from speech acoustic features.
Since gestures require time to prepare in order to be in synchrony with speech, it was necessary to provide that model with 1 second of future speech.
That article also investigates style control of the output motion, which provides another option for constraining motion diversity.}
%This is discussed in detail, which also provides style control.
%We have also on a set of and provide an example of this in the supplementary video.

%\subsection{Investigating probabilistic motion generation}
%\label{ssec:randomness}

\subsection{\new{Drawbacks and limitations}}
\label{ssec:limitations}
\new{While being a powerful machine-learning method, MoGlow comes with some disadvantages of note in computer-graphics scenarios.
Aside from the fact that machine learning affords less direct control over motion than hand animation does (and thus is more suited to high-level style control as mentioned in Sec.\ \ref{ssec:exploration}), the most relevant limitations relate to resource use at training and synthesis time.

Training a model like MoGlow demands substantial amounts of data and computation.
In many graphics applications, waiting several hours to obtain an updated model is undesirable.
Iteration time during model development may be sped up by training on multiple GPUs and by using model-surgery techniques \cite{openai2019dota} to avoid re-training new architectures from scratch.
As for data, the various training and validation curves reported in \citet{alexanderson2020robust} suggest that the MG systems in this article are ``data-limited'', and that more training data should improve held-out data likelihood.
Aside from recording additional material or pre-training on other motion databases, one might use high-quality data-augmentation techniques like those in \citet{lee2018interactive} to increase training-set size.
This can be seen as a way to inject domain knowledge into the model-creation process.
%As long as the augmentation procedure, better models are likley to resut.

%A related issue is resource use during training, both in terms of data and computation.
%Like many contemporary machine-learning systems, MoGlow relies on a large database of training material for best results.
%Results in \citet{alexanderson2020robust} suggest that the training and validation curves of MoGlow systems generally are consistent with those exhibited by Glow systems trained on small databases.
%This suggests that model accuracy has not saturated, so additional data may lead to better results, but conversely also that accuracy is likely to decrease if smaller database sizes are used.
%One interesting is to apply high-quality data augmentation methods such as those described in \citet{lee2018interactive}.

MoGlow requires that frames are generated in sequence.
Since the method describes an entire distribution of plausible poses, models furthermore tend to be deep and large.
These properties may complicate interactive applications such as games.
In general, it is easier to make good models fast than it is to make fast models good, and we expect it to be entirely possible to speed up MoGlow generation, e.g., using density distillation techniques like \citet{huang2020probability} to create shallower models with similar accuracy as deeper ones.
To compress the model footprint, neural-network pruning techniques like those surveyed in \citet{blalock2020state} are a compelling choice.}
%
%It is not unexpected that deterministic task-specific models such as QN and MA are more parsimonious than a general-purpose method that makes.
%Furthermore, MG describes an entire probability distribution of plausible pose sequences, which the deterministic state-of-the-art baselines QN and MA do not.
%Nonetheless, it is of great interest to reduce computational and memory demands for many interactive applications.
%In general, it is easier to make good models fast than it is to fast models good, and we expect it to be entirely possible to compress the model using recent advances in neural-network pruning such as those surveyed in \citet{blalock2020state}.
%
%This can .
%specifically the time .
%turnaround times are ideal for many tasks.
%
%Aside from issues of resources, machine learning also affords less direct control than hand animation.
%We saw on.
%, or by
%controllability.
%So on a diverse database like that in Sec.\ \ref{ssec:exploration}.
%This is a general trade-off. MoGlow seems.
%To achieve

\newer{While MoGlow has performed well on the various motion tasks we have tried it on, we note that it does not contain any explicit physics model.
We have seen rare instances of physically inappropriate motion, such as leaning stances where a real character would fall over.
Reverse-time augmentation, when used, can give similar issues such as leaning forwards when running backwards at speed.
We expect that these issues can be mitigated by more training data (reducing the need for augmentation), and by providing contact information as an input signal, but it might be more efficient to consider methods for introducing physics directly into the model.
MoGlow also does not contain any model of human behaviour and intent, so in the absence of external information to guide the choice of behaviour, model output may switch between diverse locomotion modes and styles in an unstructured manner.}

\section{Conclusion and future work}
\label{sec:conclusion}
We have described the first model of motion-data sequences based on normalising flows.
This paradigm is attractive because flows 1) are probabilistic (unlike many established motion models), 2) utilise powerful, implicitly defined distributions (like GANs, but unlike classical autoregressive models), yet 3) are trained to directly maximise the exact data likelihood (unlike GANs and VAEs).
%Both unconditional and conditional (i.e., controllable) models have been described.
Our model uses both autoregression and a hidden state (recurrence) to generate output sequentially, and incorporates a control scheme without algorithmic latency.
(Non-causal control is a straightforward extension.)
To our knowledge, no other Glow-based sequence models combine these desirable traits, and no other such model has incorporated hidden states, nor data dropout for more consistent control.
Moreover, our approach is probabilistic from the ground up and generates convincing samples without entropy-reduction schemes like those in \citet{greenwood2017predicting,greenwood2017predictingb,brock2019large,henter2016minimum}.
Experimental evaluations show that the model produces high-quality synthetic locomotion for both bipedal and quadrupedal motion-capture data, despite their disparate morphologies.
Subjective and objective results show that our proposal significantly outperforms task-agnostic LSTM and VAE-based approaches, coming close to natural motion recordings and performing on par with \new{task-specific state-of-the-art locomotion models}.
%In a subjective evaluation of human locomotion generation, the rated quality of randomly-sampled motions was numerically close to that of natural motion, with no significant difference between fixed-length models closer to the original Glow \cite{kingma2018glow} and our faster, simpler, sequential, and causal MoGlow model.

In light of the quality of the synthesised motion and the generally-applicable nature of the approach, we believe that models based on normalising flows can prove valuable for a wide variety of tasks incorporating motion data.
Future work includes applying the method to additional tasks and domains\new{, and making models lighter and faster for applied scenarios}.
%We are also eager to explore the unique benefits of the probabilistic nature of the model, e.g., in non-deterministic output generation, motion inpainting, and classification tasks.
Since models based on normalising flows allow exact and tractable inference, another interesting application would be to use the probabilities inferred by these models to also enable classification.

% ACM requires the acks environment to acknowledge funding sources and other support, placed right before the references section
\begin{acks}
This research was partially supported by Swedish Research Council proj.\ 2018-05409 (StyleBot), Swedish Foundation for Strategic Research contract no.\ RIT15-0107 (EACare), and by the Wallenberg AI, Autonomous Systems and Software Program (WASP) funded by the Knut and Alice Wallenberg Foundation.
\end{acks}

% Bibliography
%\bibliographystyle{ACM-Reference-Format}
%\bibliography{refs}
%\balance

 % Manually insert a pre-compiled bbl-file containing a balance command

%\appendix
%\input{appendix.tex}

\end{document}